\documentclass[10pt,twocolumn,letterpaper]{article}

\usepackage{cvpr}
\usepackage{times}
\usepackage{epsfig}
\usepackage{graphicx}
\usepackage{amsmath}
\usepackage{amssymb}
\usepackage{color}
\usepackage{times}
\usepackage{epsfig}
\usepackage{algorithm}
\usepackage{algorithmicx}
\usepackage{algpseudocode}
\usepackage{graphics}
\usepackage{threeparttable}
\usepackage{color}
\usepackage[normalem]{ulem}
\usepackage{multirow}
\usepackage{float}
\usepackage{amsfonts}
\usepackage{bm}
\usepackage{subfig}
\usepackage{array}
\usepackage[table]{xcolor}
\usepackage{colortbl}
\usepackage{pifont}
\usepackage{cite}
\usepackage{diagbox}
\usepackage{rotating}
\usepackage{booktabs}
\usepackage{overpic}
\usepackage{textcomp}
\usepackage{contour}
\usepackage{enumitem}

\usepackage[pagebackref=true,breaklinks=true,letterpaper=true,colorlinks,bookmarks=false]{hyperref}
\cvprfinalcopy 

\def\cvprPaperID{****} 

\ifcvprfinal\fi
\begin{document}

\title{Zero-Reference Deep Curve Estimation for Low-Light Image Enhancement}

\author{Chunle Guo$^1$$^,$$^2$\footnotemark[1]\quad  Chongyi Li$^1$$^,$$^2$\footnotemark[1]\quad  Jichang Guo$^1$\footnotemark[2]\quad  \\
	\quad Chen Change Loy$^3$\quad  Junhui Hou$^2$\quad Sam Kwong$^2$\quad Runmin Cong$^4$\\
	\small{$^1$} \small BIIT Lab, Tianjin University \quad
	\small{$^2$} \small City University of Hong Kong \quad
	\small{$^3$} \small Nanyang Technological University\quad
	\small{$^4$} \small Beijing Jiaotong University \quad\\
	{\tt\small $\{$guochunle,lichongyi,jcguo$\}$@tju.edu.cn  \quad ccloy@ntu.edu.sg}\\
		{\tt\small$\{$jh.hou,cssamk$\}$@cityu.edu.hk \quad rmcong@bjtu.edu.cn }\\
	{\tt\small \url{https://li-chongyi.github.io/Proj_Zero-DCE.html/}}
}

\maketitle
\thispagestyle{empty}
\renewcommand{\thefootnote}{\fnsymbol{footnote}}
\footnotetext[1]{The first two authors contribute equally to this work.}
\footnotetext[2]{Jichang Guo (jcguo@tju.edu.cn) is the corresponding author.}
\vspace{-8pt}
\begin{abstract}
The paper presents a novel method, Zero-Reference Deep Curve Estimation (Zero-DCE), which formulates light enhancement as a task of image-specific curve estimation with a deep network.  Our method trains a lightweight deep network, DCE-Net, to estimate pixel-wise and high-order curves for dynamic range adjustment of a given image. The curve estimation is specially designed, considering pixel value range, monotonicity, and differentiability. Zero-DCE is appealing in its relaxed assumption on reference images, i.e., it does not require any paired or unpaired data during training. This is achieved through a set of carefully formulated non-reference loss functions, which implicitly measure the enhancement quality and drive the learning of the network. Our method is efficient as image enhancement can be achieved by an intuitive and simple nonlinear curve mapping. Despite its simplicity, we show that it generalizes well to diverse lighting conditions. Extensive experiments on various benchmarks demonstrate the advantages of our method over state-of-the-art methods qualitatively and quantitatively. Furthermore, the potential benefits of our Zero-DCE to face detection in the dark are discussed.
\end{abstract}

\vspace{-6pt}
\section{Introduction}
Many photos are often captured under suboptimal lighting conditions due to inevitable environmental and/or technical constraints. These include inadequate and unbalanced lighting conditions in the environment, incorrect placement of objects against extreme back light, and under-exposure during image capturing. Such low-light photos suffer from compromised aesthetic quality and unsatisfactory transmission of information. The former affects viewers' experience while the latter leads to wrong message being communicated, such as inaccurate object/face recognition.


\begin{figure}
	\begin{center}
		\begin{tabular}{c@{ }c@{ }c@{ }c@{ }c}
			\includegraphics[height=3cm]{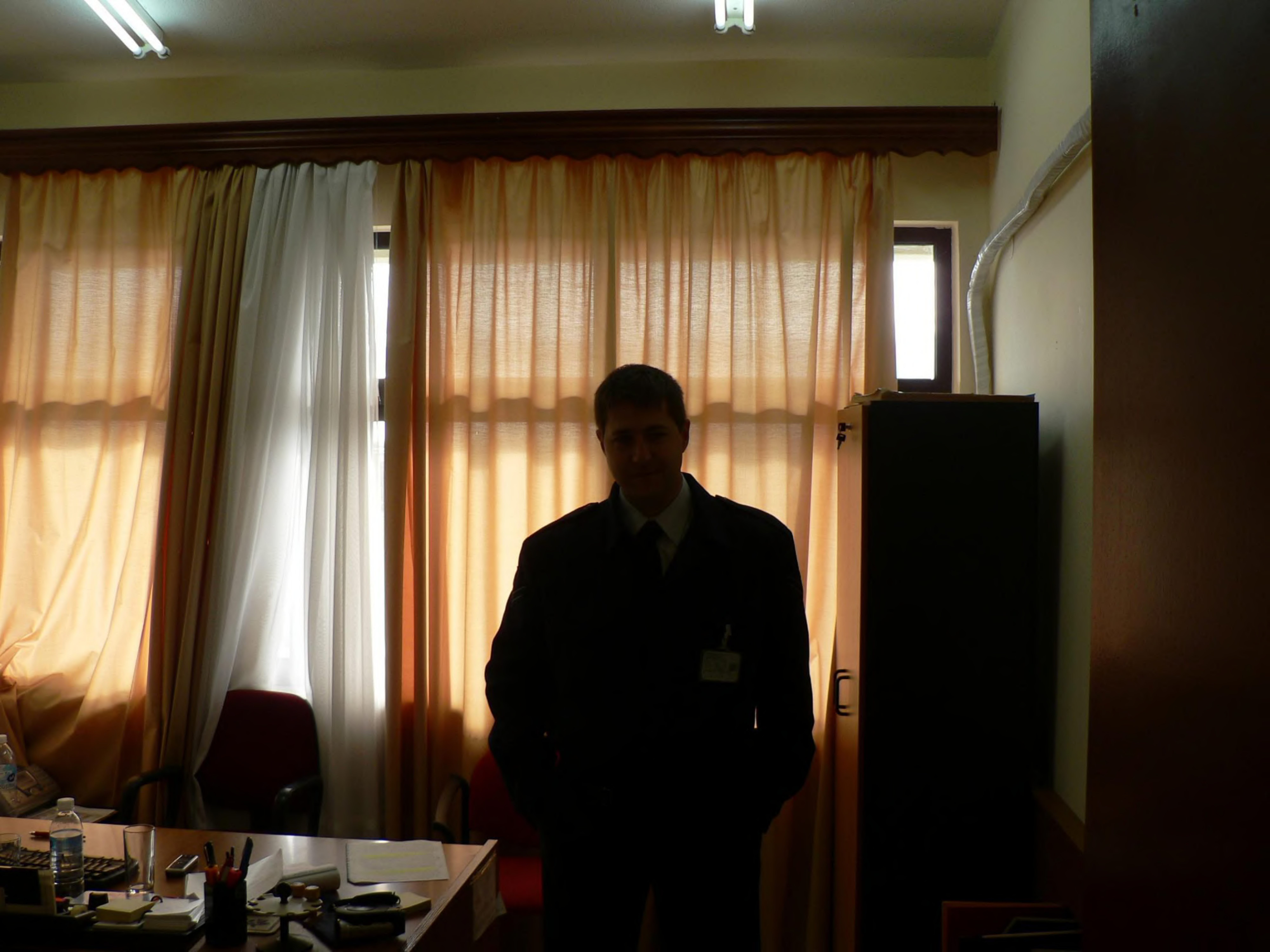}&
			\includegraphics[height=3cm]{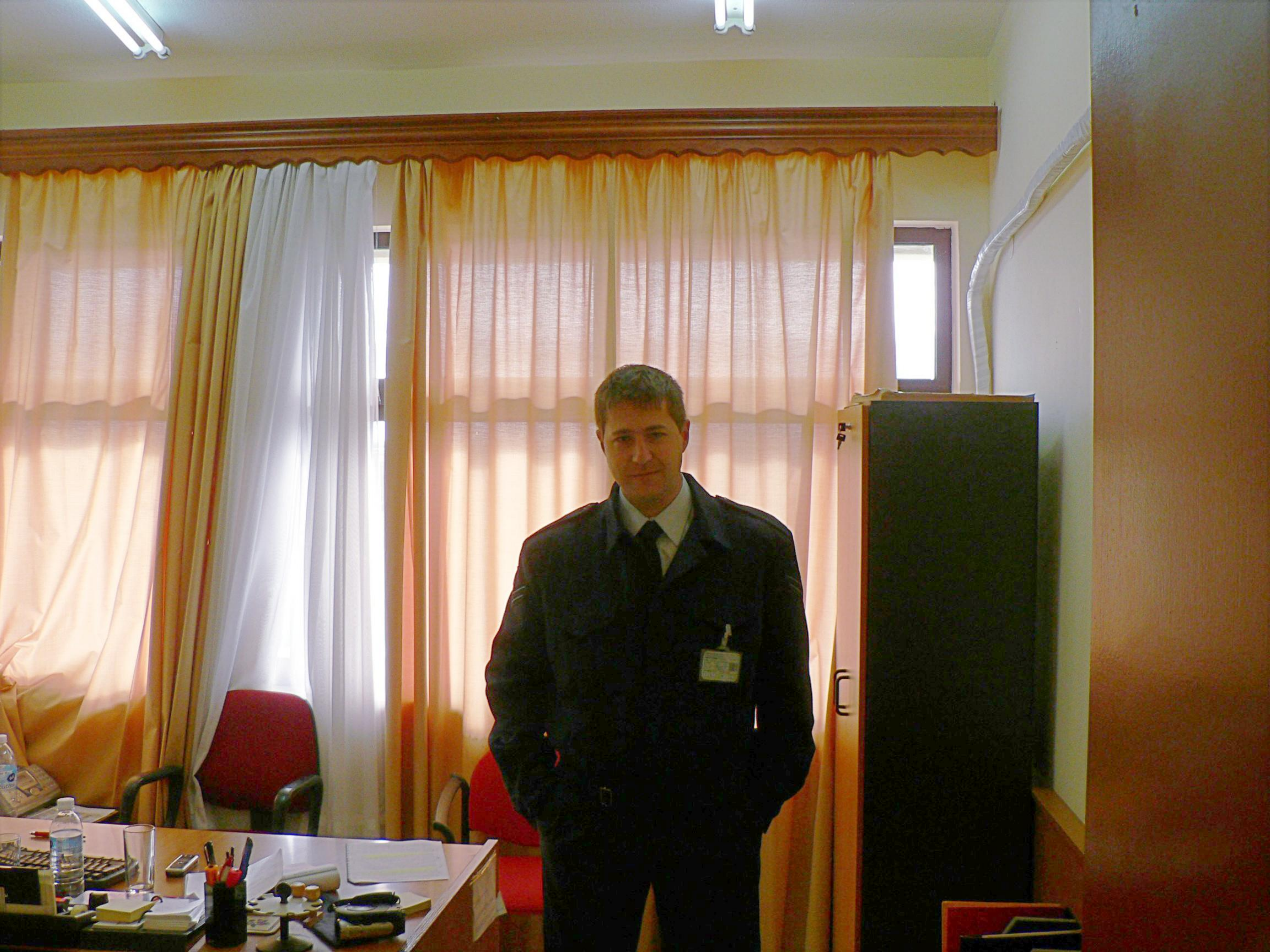}\\
			(a) Raw & (b) Zero-DCE \\
			\includegraphics[height=3cm]{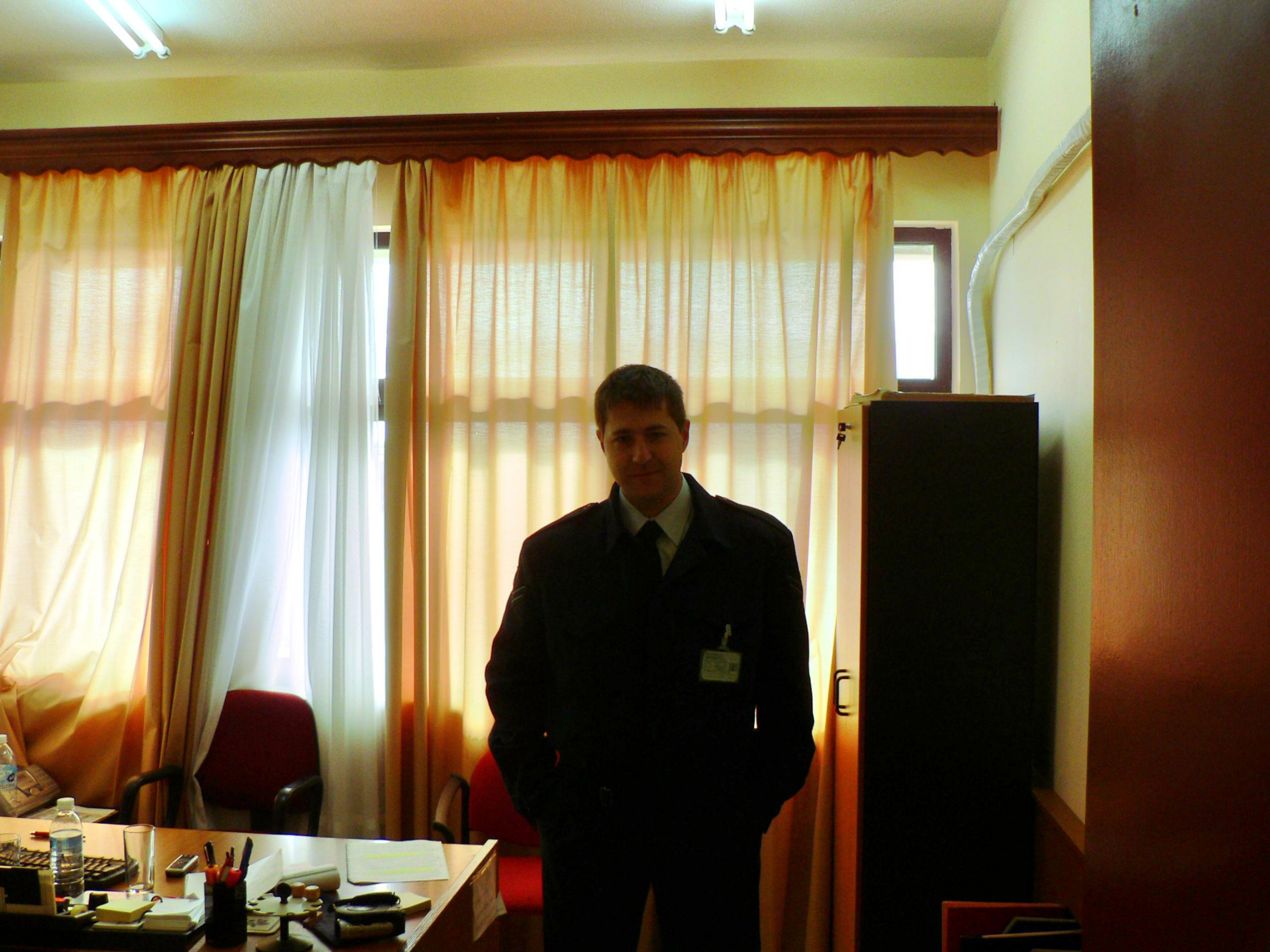}&
			\includegraphics[height=3cm]{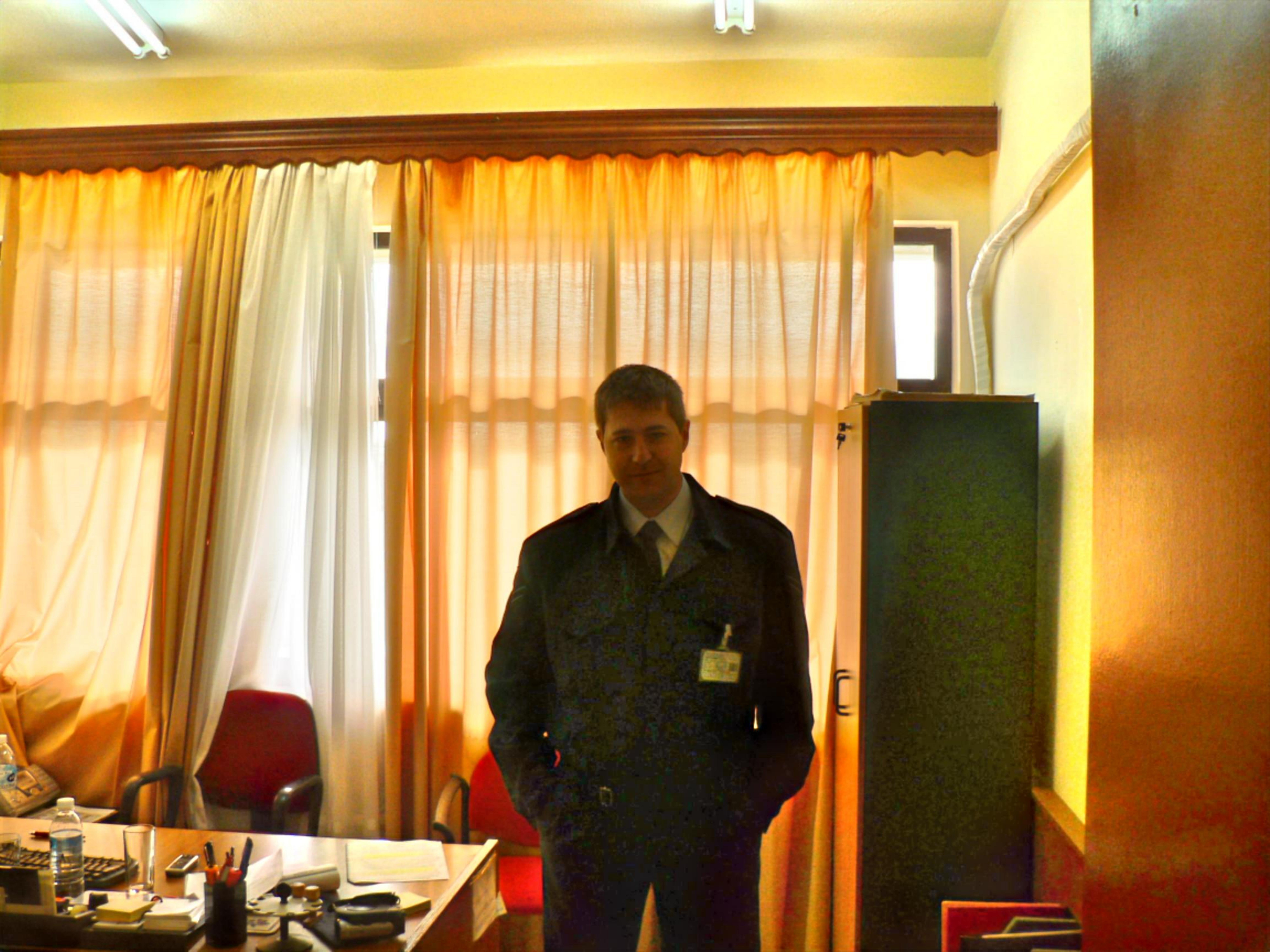}\\
			(c) Wang \etal~\cite{Wang2019} & (d) EnlightenGAN~\cite{Jiang2019}\\
		\end{tabular}
	\end{center}
	\vspace{-0.5cm}
	\caption{Visual comparisons on a typical low-light image. The proposed Zero-DCE achieves visually pleasing result in terms of brightness, color, contrast, and naturalness, while existing methods either fail to cope with the extreme back light or generate color artifacts. In contrast to other deep learning-based methods, our approach is trained without any reference image.}
	\label{fig:im_sample2}
	\vspace{-0.5cm}
\end{figure}

%


In this study, we present a novel deep learning-based method, Zero-Reference Deep Curve Estimation (Zero-DCE), for low-light image enhancement. It can cope with diverse lighting conditions including nonuniform and poor lighting cases.
Instead of performing image-to-image mapping, we reformulate the task as an image-specific curve estimation problem.
In particular, the proposed method takes a low-light image as input and produces high-order curves as its output. These curves are then used for pixel-wise adjustment on the dynamic range of the input to obtain an enhanced image.
The curve estimation is carefully formulated so that it maintains the range of the enhanced image and preserves the contrast of neighboring pixels. Importantly, it is differentiable, and thus we can learn the adjustable parameters of the curves through a deep convolutional neural network.
The proposed network is lightweight and it can be iteratively applied to approximate higher-order curves for more robust and accurate dynamic range adjustment.

A unique advantage of our deep learning-based method is \textbf{zero-reference}, \ie, it does not require any paired or even unpaired data in the training process as in existing CNN-based~\cite{Wang2019, Chen2018} and GAN-based methods~\cite{Jiang2019,CycleGAN}.
This is made possible through a set of specially designed non-reference loss functions including spatial consistency loss, exposure control loss, color constancy loss, and illumination smoothness loss, all of which take into consideration multi-factor of light enhancement.
We show that even with zero-reference training, Zero-DCE can still perform competitively against other methods that require paired or unpaired data for training.
An example of enhancing a low-light image comprising nonuniform illumination is shown in Fig.~\ref{fig:im_sample2}.
Comparing to state-of-the-art methods, Zero-DCE brightens up the image while preserving the inherent color and details. In contrast, both CNN-based method~\cite{Wang2019} and GAN-based EnlightenGAN~\cite{Jiang2019} yield under-(the face) and over-(the cabinet) enhancement.


Our \textbf{contributions} are summarized as follows.

\noindent 1)
We propose the first low-light enhancement network that is independent of paired and unpaired training data, thus avoiding the risk of overfitting. As a result, our method generalizes well to various lighting conditions.

\noindent 2)
We design an image-specific curve that is able to approximate pixel-wise and higher-order curves by iteratively applying itself. Such image-specific curve can effectively perform mapping within a wide dynamic range.

\noindent 3)
We show the potential of training a deep image enhancement model in the absence of reference images through task-specific non-reference loss functions that indirectly evaluate enhancement quality.

Our Zero-DCE method supersedes state-of-the-art performance both in qualitative and quantitative metrics. More importantly, it is capable of improving high-level visual tasks, \eg, face detection, without inflicting high computational burden. \textbf{It is capable of processing images in real-time (about 500 FPS for images of size 640$\times$480$\times$3 on GPU) and takes only 30 minutes for training.}

\vspace{-5pt}
\section{Related Work}

\noindent
\textbf{Conventional Methods.}
HE-based methods perform light enhancement through expanding the dynamic range of an image. Histogram distribution of images is adjusted at both global~\cite{Coltuc2006,Ibrahim2007} and local levels~\cite{Stark2000,Lee2013}. There are also various methods adopting the Retinex theory~\cite{Land1977} that typically decomposes an image into reflectance and illumination. The reflectance component is commonly assumed to be consistent under any lighting conditions; thus, light enhancement is formulated as an illumination estimation problem. Building on the Retinex theory, several methods have been proposed. Wang \etal~\cite{Wang2013} designed a naturalness- and information-preserving method when handling images of nonuniform illumination; Fu \etal~\cite{Fu2016} proposed a weighted variation model to simultaneously estimate the reflectance and illumination of an input image; Guo \etal~\cite{Guo2017} first estimated a coarse illumination map by searching the maximum intensity of each pixel in RGB channels, then refining the coarse illumination map by a structure prior; Li \etal~\cite{Li2018} proposed a new Retinex model that takes noise into consideration. The illumination map was estimated through solving an optimization problem.
Contrary to the conventional methods that fortuitously change the distribution of image histogram or that rely on potentially inaccurate physical models, the proposed Zero-DCE method produces an enhanced result through image-specific curve mapping. Such a strategy enables light enhancement on images without creating unrealistic artifacts.
Yuan and Sun \cite{Yuan2012} proposed an automatic exposure correction method, where the S-shaped curve for a given image is estimated by a global optimization algorithm and each segmented region is pushed to its optimal zone by curve mapping. Different from \cite{Yuan2012}, our Zero-DCE is a purely data-driven method and takes multiple light enhancement factors into consideration in the design of the non-reference loss functions, and thus enjoys better robustness, wider image dynamic range adjustment, and lower computational burden.

\noindent
\textbf{Data-Driven Methods.}
Data-driven methods are largely categorized into two branches, namely CNN-based and GAN-based methods. Most CNN-based solutions rely on paired data for supervised training, therefore they are resource-intensive. Often time, the paired data are exhaustively collected through automatic light degradation, changing the settings of cameras during data capturing, or synthesizing data via image retouching. For example, the LL-Net~\cite{Lore2017} was trained on data simulated on random Gamma correction; the LOL dataset~\cite{Chen2018} of paired low/normal light images was collected through altering the exposure time and ISO during image acquisition; the MIT-Adobe FiveK dataset~\cite{Adobe5K} comprises 5,000 raw images, each of which has five retouched images produced by trained experts. 
\begin{figure*}[t]
	\centering
	\centerline{\includegraphics[width=0.92\linewidth]{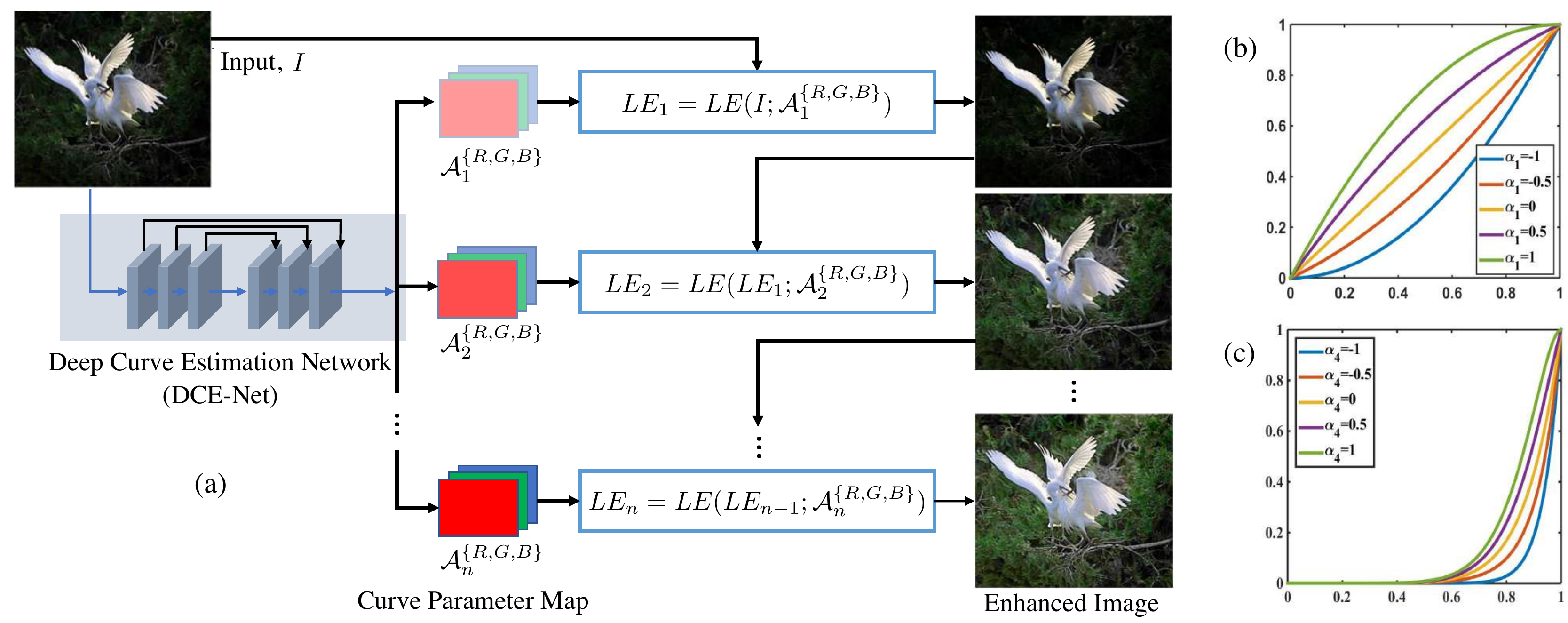}}
	\vspace{-0.4cm}
	\caption{(a) The framework of Zero-DCE. A DCE-Net is devised to estimate a set of best-fitting Light-Enhancement curves (LE-curves) that iteratively enhance a given input image. (b, c) LE-curves with different adjustment parameters $\alpha$ and numbers of iteration $n$. In (c), $\alpha_{1}$, $\alpha_{2}$, and $\alpha_{3}$ are equal to -1 while $n$ is equal to 4. In each subfigure, the horizontal axis represents the input pixel values while the vertical axis represents the output pixel values. }
	\label{fig:pipeline}
	\vspace{-0.4cm}
\end{figure*}

Recently, Wang \etal~\cite{Wang2019} proposed an underexposed photo enhancement network by estimating the illumination map. This network was trained on paired data that were retouched by three experts.
Understandably, light enhancement solutions based on paired data are impractical in many ways, considering the high cost involved in collecting sufficient paired data as well as the inclusion of factitious and unrealistic data in training the deep models. Such constraints are reflected in the poor generalization capability of CNN-based methods. Artifacts and color casts are commonly generated, when these methods are presented with real-world images of various light intensities.

Unsupervised GAN-based methods have the advantage of eliminating paired data for training. EnlightenGAN~\cite{Jiang2019}, an unsupervised GAN-based and pioneer method that learns to enhance low-light images using unpaired low/normal light data. The network was trained by taking into account elaborately designed discriminators and loss functions. However, unsupervised GAN-based solutions usually require careful selection of unpaired training data.

The proposed Zero-DCE is superior to existing data-driven methods in three aspects. First, it explores a new learning strategy, \ie, one that requires \textit{zero reference}, hence eliminating the need for paired and unpaired data. Second, the network is trained by taking into account carefully defined non-reference loss functions. This strategy allows output image quality to be implicitly evaluated, the results of which would be reiterated for network learning. Third, our method is highly efficient and cost-effective. These advantages benefit from our zero-reference learning framework, lightweight network structure, and effective non-reference loss functions.

\vspace{-5pt}
\section{Methodology}

We present the framework of Zero-DCE in Fig.~\ref{fig:pipeline}.
A Deep Curve Estimation Network (DCE-Net) is devised to estimate a set of best-fitting Light-Enhancement curves (LE-curves) given an input image.
The framework then maps all pixels of the input's RGB channels by applying the curves iteratively for obtaining the final enhanced image.
We next detail the key components in Zero-DCE, namely LE-curve, DCE-Net, and non-reference loss functions in the following sections.

\vspace{-5pt}
\subsection{Light-Enhancement Curve (LE-curve)}

Inspired by the curves adjustment used in photo editing software, we attempt to design a kind of curve that can map a low-light image to its enhanced version automatically, where the self-adaptive curve parameters are solely dependent on the input image.
There are three objectives in the design of such a curve: 1) each pixel value of the enhanced image should be in the normalized range of [0,1] to avoid information loss induced by overflow truncation; 2) this curve should be monotonous to preserve the differences (contrast) of neighboring pixels; and 3) the form of this curve should be as simple as possible and differentiable in the process of gradient backpropagation.

%

To achieve these three objectives, we design a quadratic curve, which can be expressed as:
\begin{equation}
\label{equ_1}
LE(I(\mathbf{x});\alpha)=I(\mathbf{x})+\alpha I(\mathbf{x})(1-I(\mathbf{x})),
\end{equation}
where $\mathbf{x}$ denotes pixel coordinates, $LE(I(\mathbf{x});\alpha)$ is the enhanced version of the given input $I(\mathbf{x})$, $\alpha\in[-1,1]$ is the trainable curve parameter, which adjusts the magnitude of LE-curve and also controls the exposure level. Each pixel is normalized to $[0,1]$ and all operations are pixel-wise.
We separately apply the LE-curve to three RGB channels instead of solely on the illumination channel. The three-channel adjustment can better preserve the inherent color and reduce the risk of over-saturation. We report more details in the supplementary material.
%

An illustration of LE-curves with different adjustment parameters $\alpha$ is shown in Fig.~\ref{fig:pipeline}(b).
It is clear that the LE-curve complies with the three aforementioned objectives.
In addition, the LE-curve enables us to increase or decrease the dynamic range of an input image. This capability is conducive to not only enhancing low-light regions but also removing over-exposure artifacts.
%
%

\begin{figure*}
	\begin{center}
		\begin{tabular}{c@{ }c@{ }c@{ }c@{ }c@{ }c}
			\includegraphics[width=.15\textwidth,height=2.55cm]{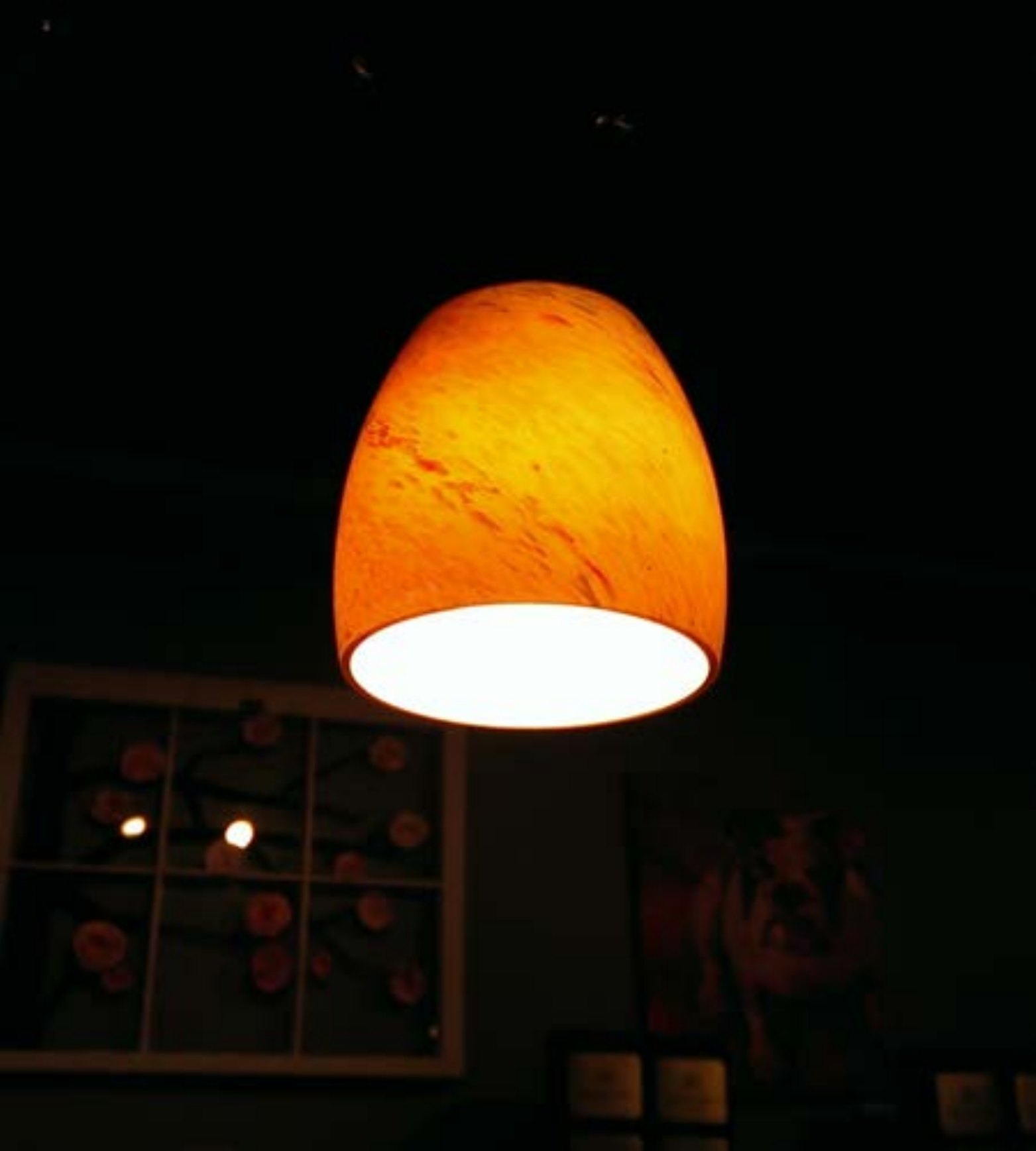}~~~~~~~&
			\includegraphics[width=.165\textwidth,height=2.65cm]{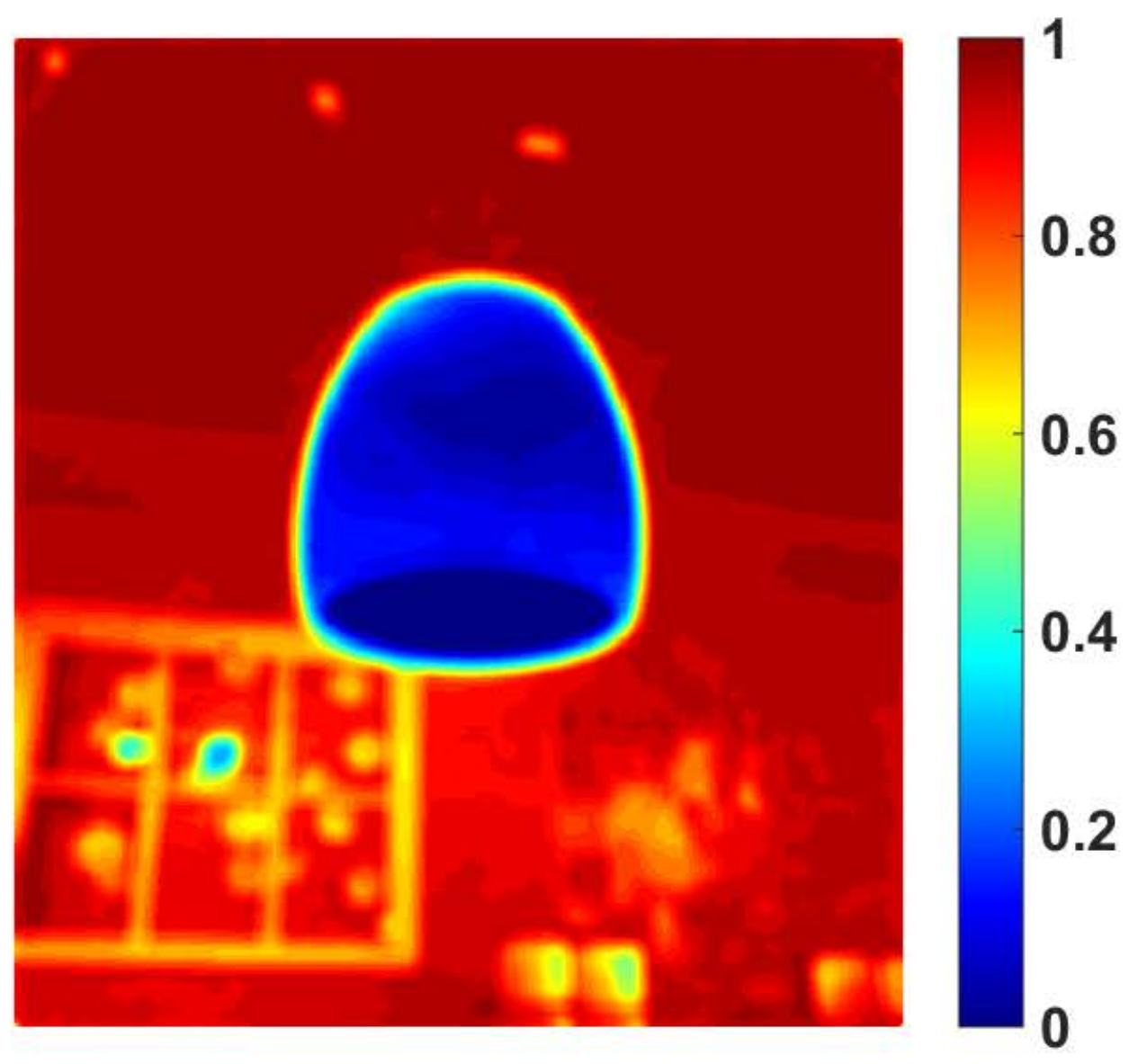}~~~~~~~&
			\includegraphics[width=.15\textwidth,height=2.6cm]{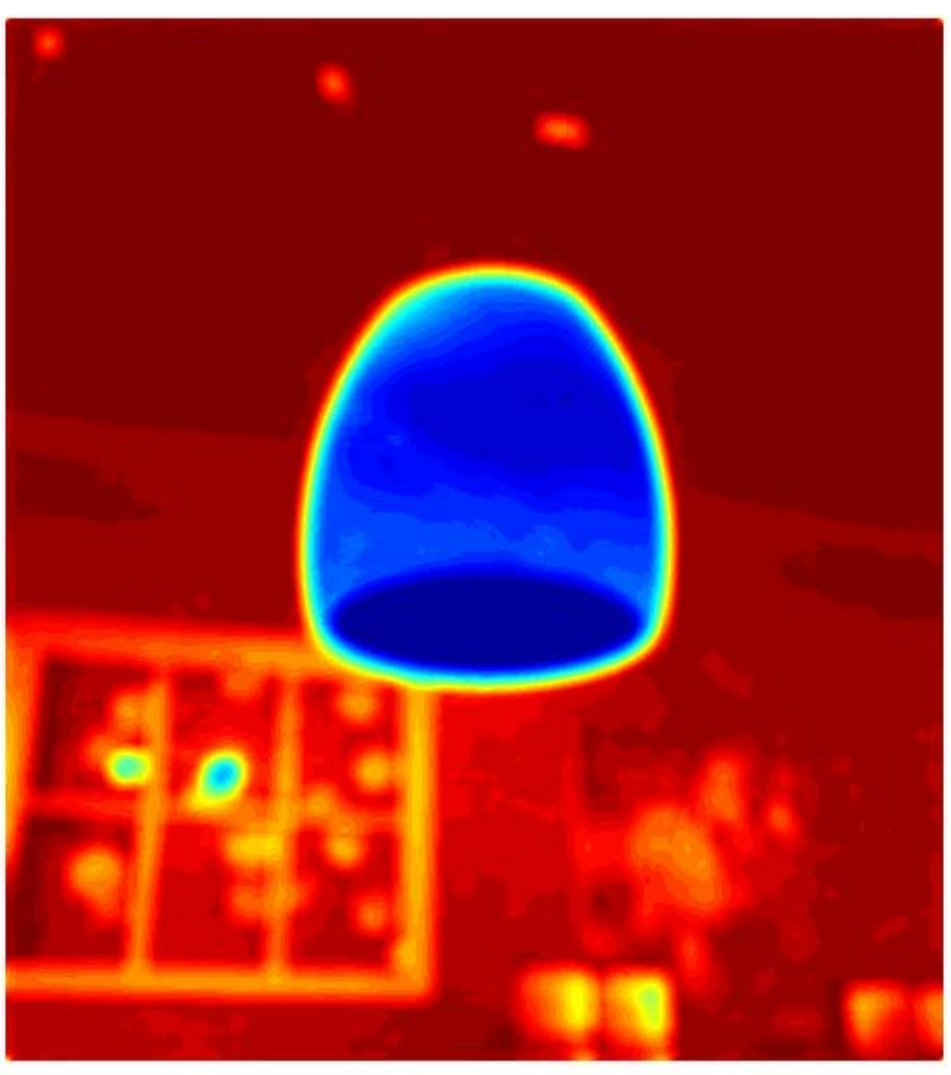}~~~~~~~&
			\includegraphics[width=.15\textwidth,height=2.6cm]{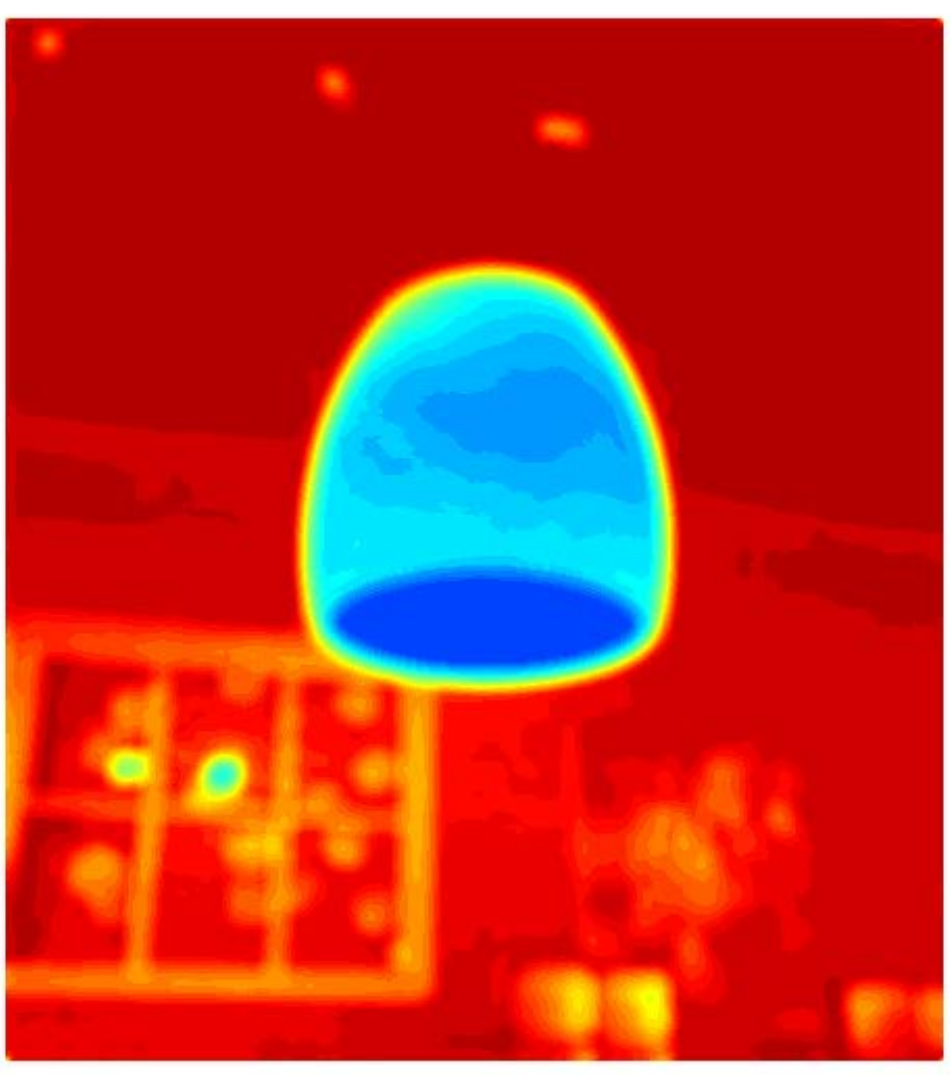}~~~~~~~&
			\includegraphics[width=.15\textwidth,height=2.55cm]{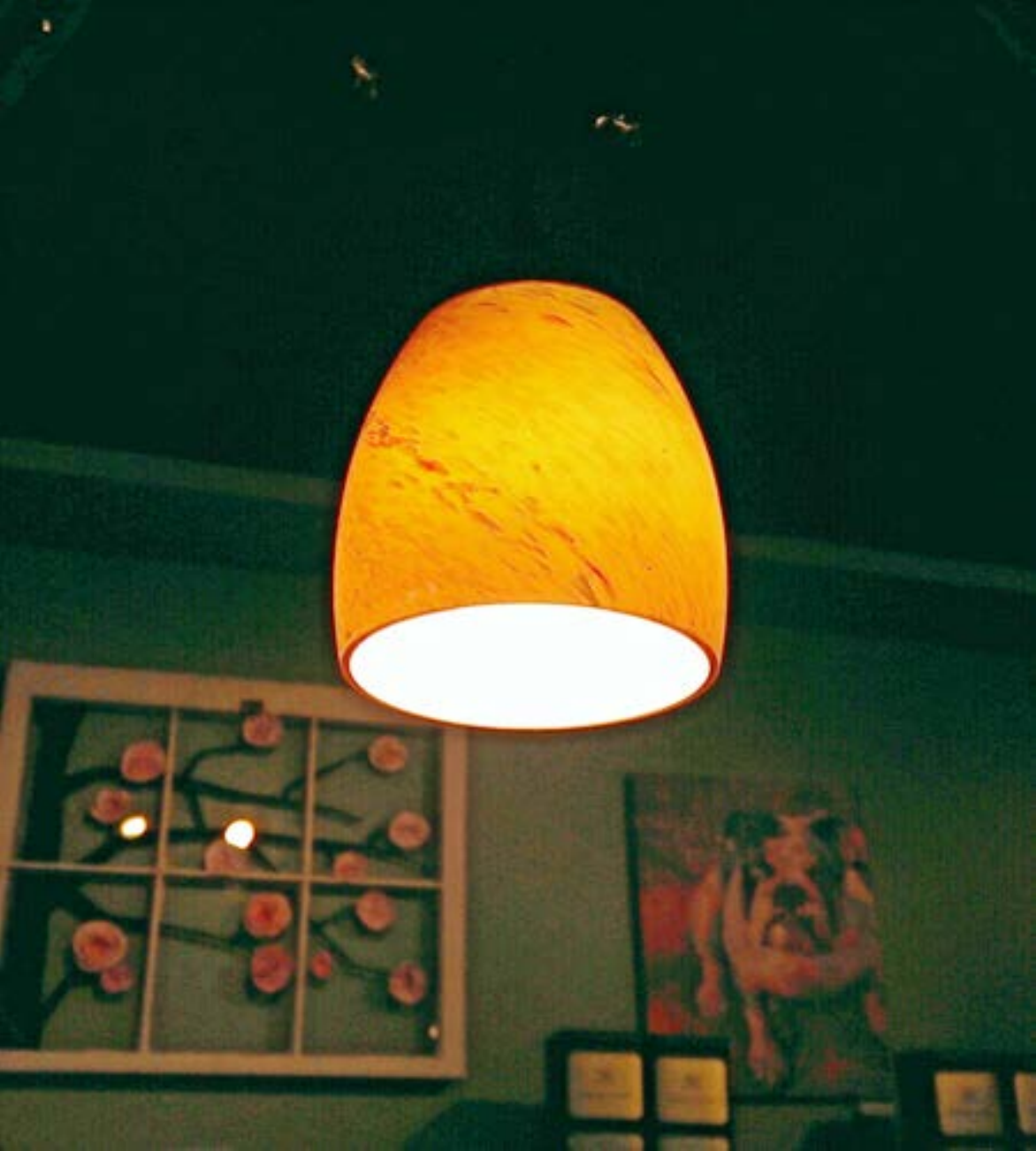}\\
			(a) Input~~~~~~& (b) $\mathcal{A}_{n}^{R}$~~~~~~& (c) $\mathcal{A}_{n}^{G}$~~~~~~& (d) $\mathcal{A}_{n}^{B}$~~~~~~& (e) Result \\
		\end{tabular}
	\end{center}
	\vspace{-0.5cm}
	\caption{An example of the pixel-wise curve parameter maps. For visualization, we average the curve parameter maps of all iterations ($n=8$) and normalize the values to the range of $[0,1]$. $\mathcal{A}_{n}^{R}$, $\mathcal{A}_{n}^{G}$, and $\mathcal{A}_{n}^{B}$ represent the averaged best-fitting curve parameter maps of R, G, and B channels, respectively. The maps in (b), (c), and (d) are represented by heatmaps.}
	\label{fig:map}
	\vspace{-0.2cm}
\end{figure*}

\noindent
\textbf{Higher-Order Curve.}
The LE-curve defined in Eq.~\eqref{equ_1} can be applied iteratively to enable more versatile adjustment to cope with challenging low-light conditions. Specifically,
%
\begin{equation}
\label{equ_2}
LE_{n}(\mathbf{x})=LE_{n-1}(\mathbf{x})+\alpha_{n}LE_{n-1}(\mathbf{x})(1-LE_{n-1}(\mathbf{x})),
\end{equation}
where $n$ is the number of iteration, which controls the curvature. In this paper, we set the value of $n$ to 8, which can deal with most cases satisfactory. Eq.~\eqref{equ_2} can be degraded to Eq.~\eqref{equ_1} when $n$ is equal to 1. Figure~\ref{fig:pipeline}(c) provides an example showing high-order curves with different $\alpha$ and $n$, which have more powerful adjustment capability (\ie, greater curvature) than the curves in Figure~\ref{fig:pipeline}(b).
%

\noindent
\textbf{Pixel-Wise Curve.}
A higher-order curve can adjust an image within a wider dynamic range. Nonetheless, it is still a global adjustment since $\alpha$ is used for all pixels.
A global mapping tends to over-/under- enhance local regions.
To address this problem, we formulate $\alpha$ as a pixel-wise parameter, \ie, each pixel of the given input image has a corresponding curve with the best-fitting $\alpha$ to adjust its dynamic range.
Hence, Eq.~\eqref{equ_2} can be reformulated as:
\begin{equation}
\label{equ_3}
LE_{n}(\mathbf{x})=LE_{n-1}(\mathbf{x})+\mathcal{A}_{n}(\mathbf{x})LE_{n-1}(\mathbf{x})(1-LE_{n-1}(\mathbf{x})),
\end{equation}
where $\mathcal{A}$ is a parameter map with the same size as the given image.
Here, we assume that pixels in a local region have the same intensity (also the same adjustment curves), and thus the neighboring pixels in the output result still preserve the monotonous relations.
In this way, the pixel-wise higher-order curves also comply with three objectives.

We present an example of the estimated curve parameter maps of three channels in Fig.~\ref{fig:map}.
As shown, the best-fitting parameter maps of different channels have similar adjustment tendency but different values, indicating the relevance and difference among the three channels of a low-light image.
The curve parameter map accurately indicates the brightness of different regions (\eg, the two glitters on the wall).
With the fitting maps, the enhanced version image can be directly obtained by pixel-wise
curve mapping. As shown in Fig.~\ref{fig:map}(e), the enhanced version reveals the content in dark regions and preserves the bright regions.

\vspace{-5pt}
\subsection{DCE-Net}
To learn the mapping between an input image and its best-fitting curve parameter maps, we propose a Deep Curve Estimation Network (DCE-Net).
The input to the DCE-Net is a low-light image while the outputs are a set of pixel-wise curve parameter maps for corresponding higher-order curves.
%
%
We employ a plain CNN of seven convolutional layers with symmetrical concatenation. Each layer consists of 32 convolutional kernels of size 3$\times$3 and stride 1 followed by the ReLU activation function. We discard the down-sampling and batch normalization layers that break the relations of neighboring pixels.
The last convolutional layer is followed by the Tanh activation function, which produces 24 parameter maps for 8 iterations ($n=8$), where each iteration requires three curve parameter maps for the three channels.
The detailed architecture of DCE-Net is provided in the supplementary material.
It is noteworthy that DCE-Net only has 79,416 trainable parameters and 5.21G Flops for an input image of size 256$\times$256$\times$3. It is therefore lightweight and can be used in computational resource-limited devices, such as mobile platforms.

\begin{figure*}
	\begin{center}
		\begin{tabular}{c@{ }c@{ }c@{ }c@{ }c@{ }c@{ }c}
			\includegraphics[width=.14\textwidth,height=2.6cm]{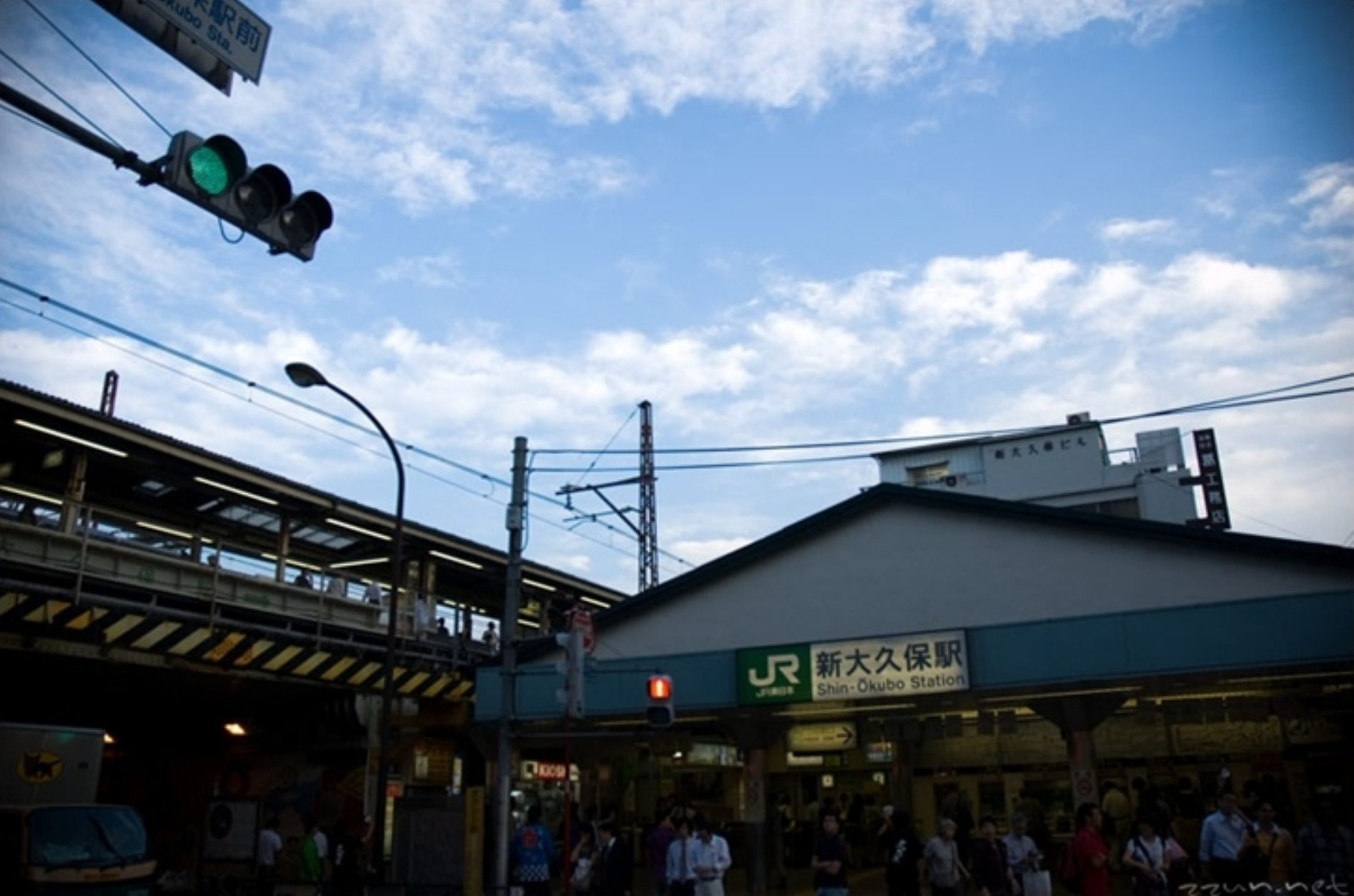}~~~~~&
			\includegraphics[width=.14\textwidth,height=2.6cm]{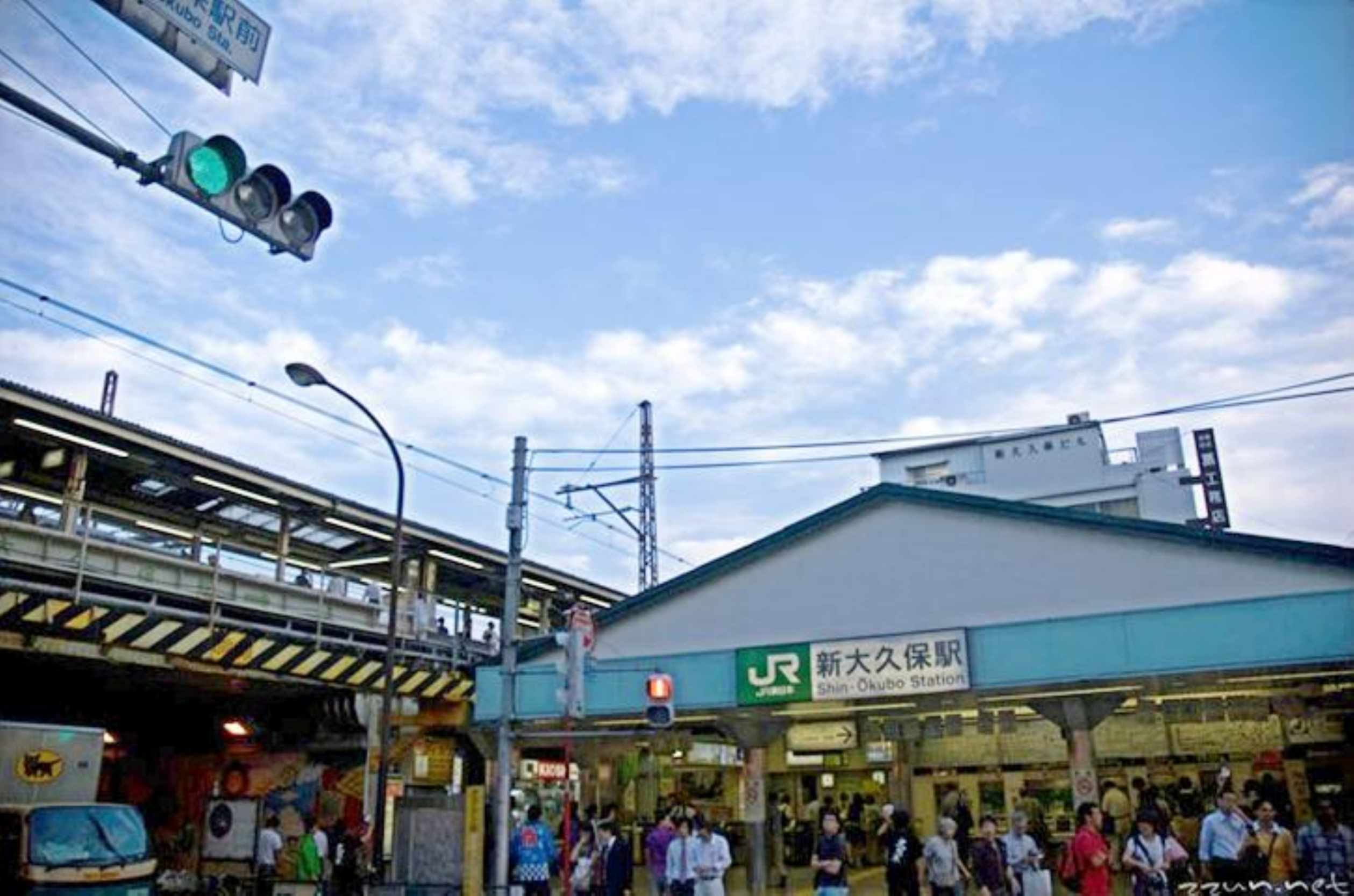}~~~~~&
			\includegraphics[width=.14\textwidth,height=2.6cm]{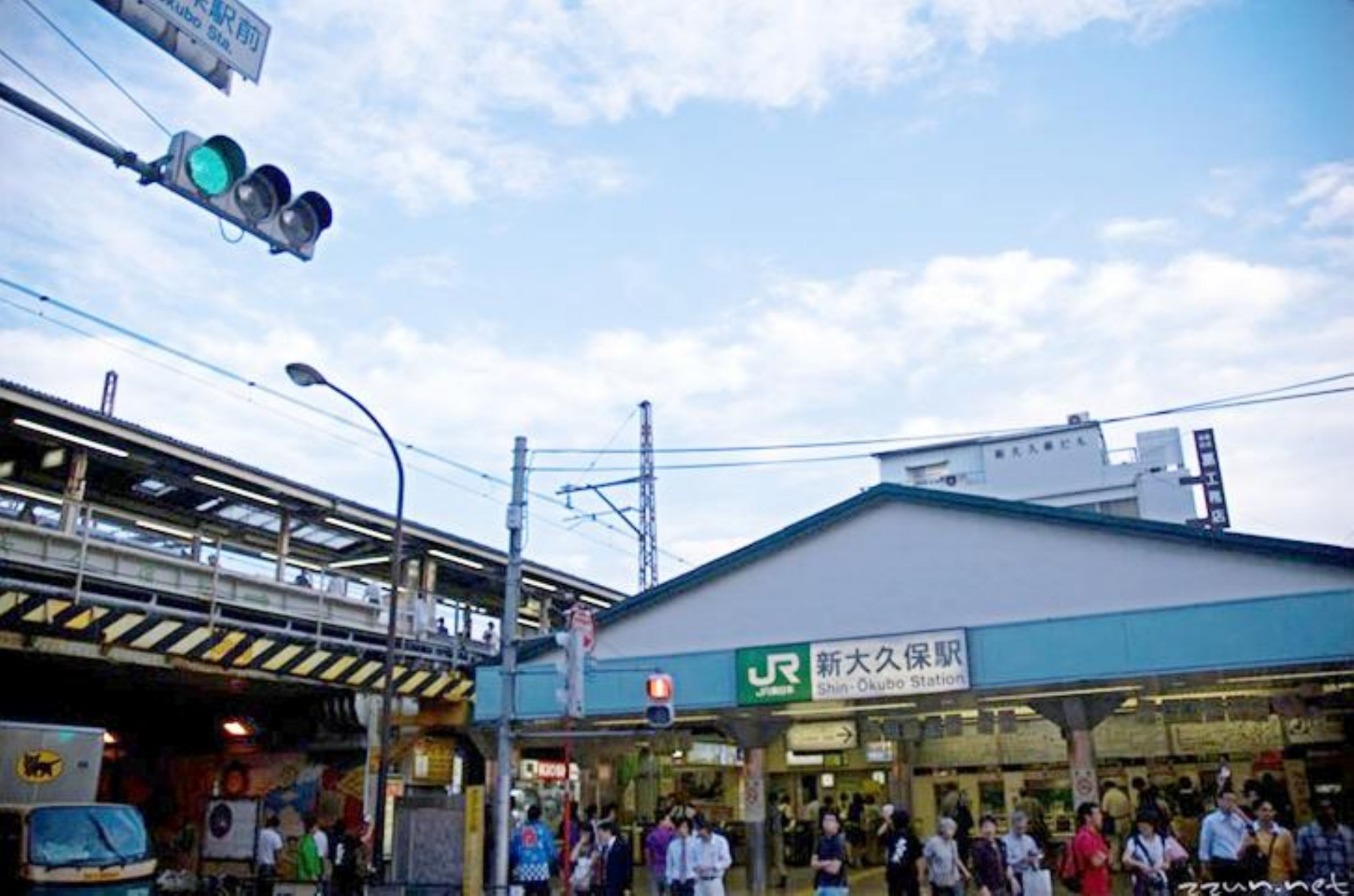}~~~~~&
			\includegraphics[width=.14\textwidth,height=2.6cm]{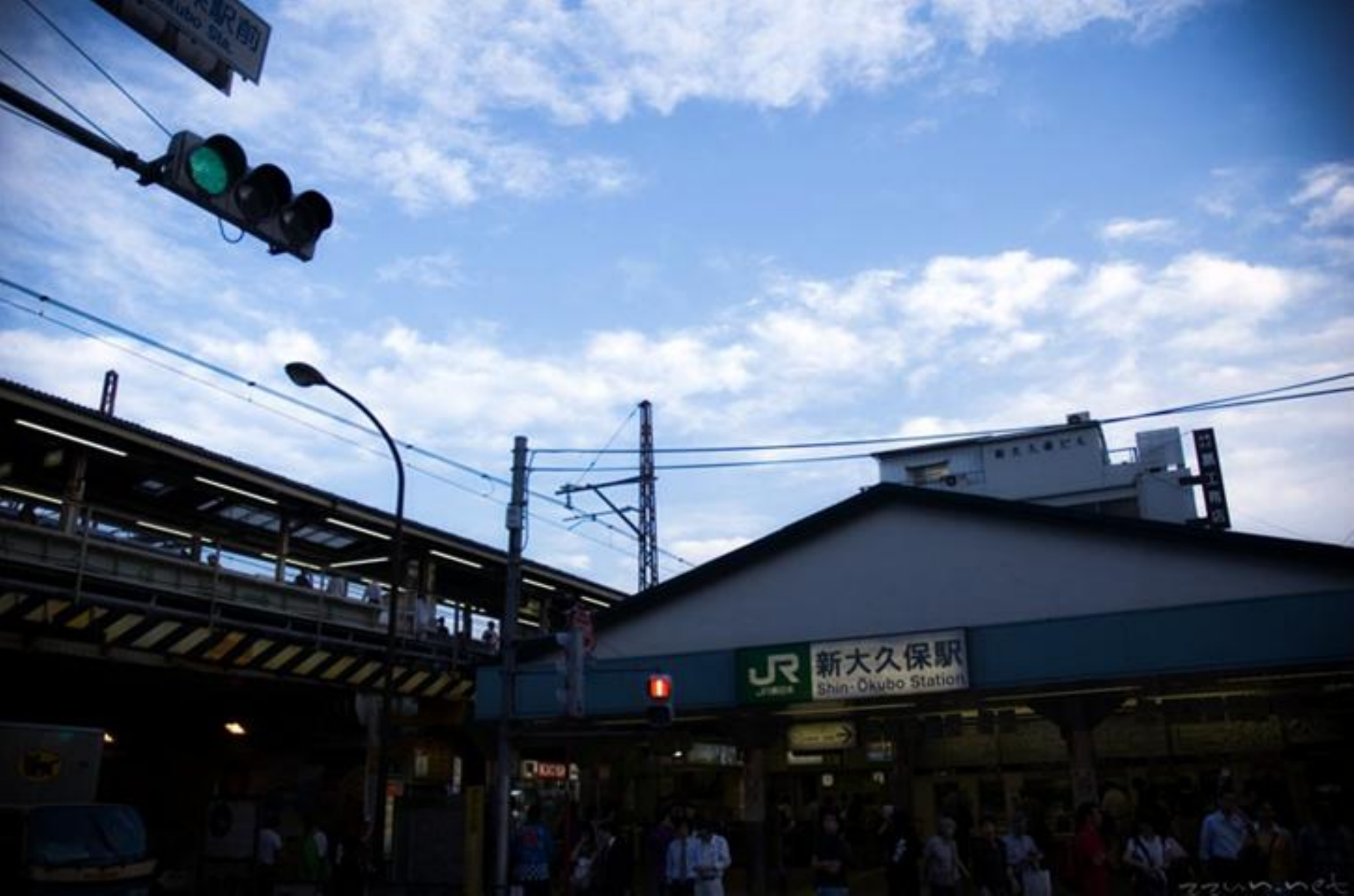}~~~~~&
			\includegraphics[width=.14\textwidth,height=2.6cm]{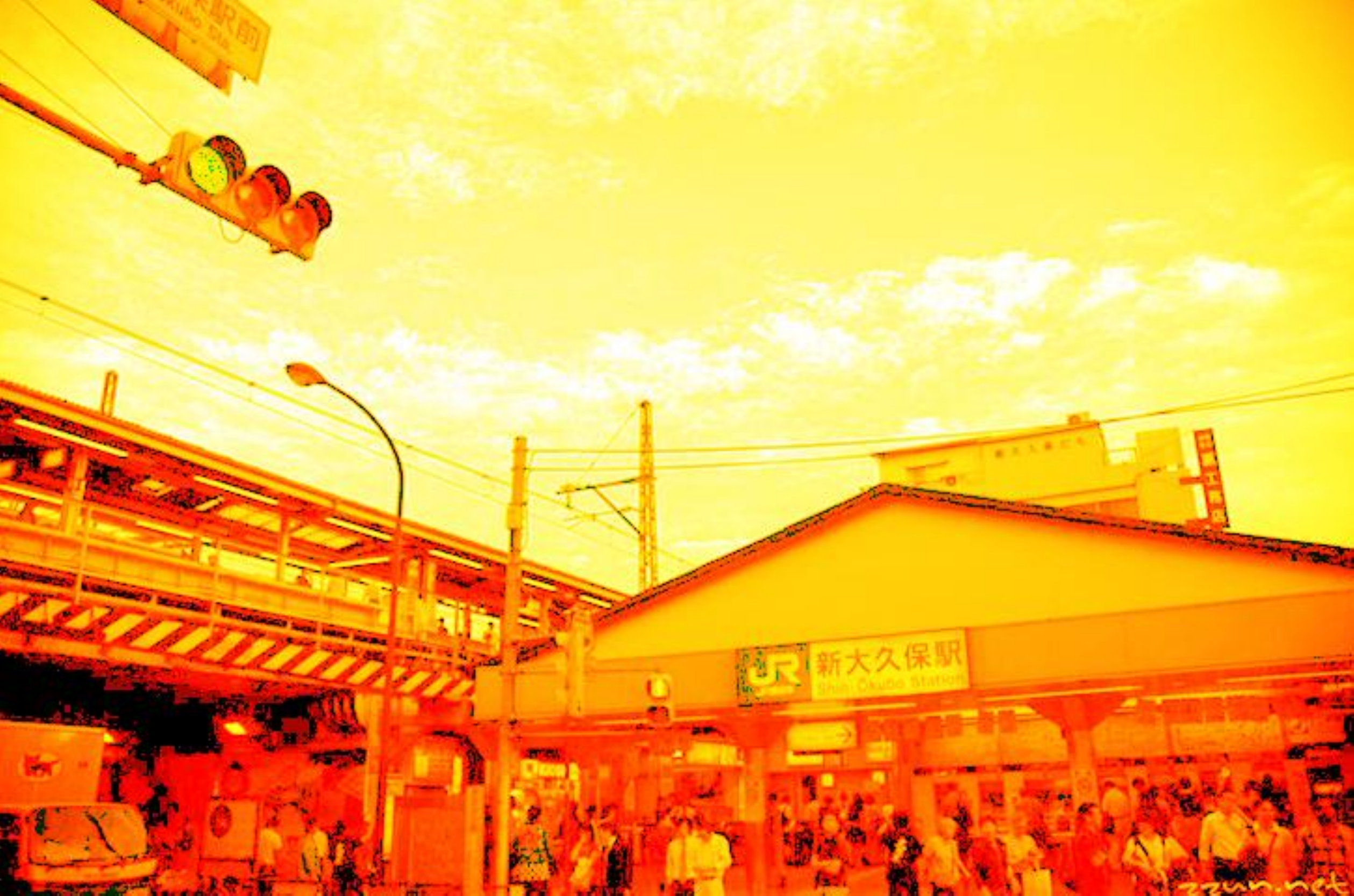}~~~~~&
			\includegraphics[width=.14\textwidth,height=2.6cm]{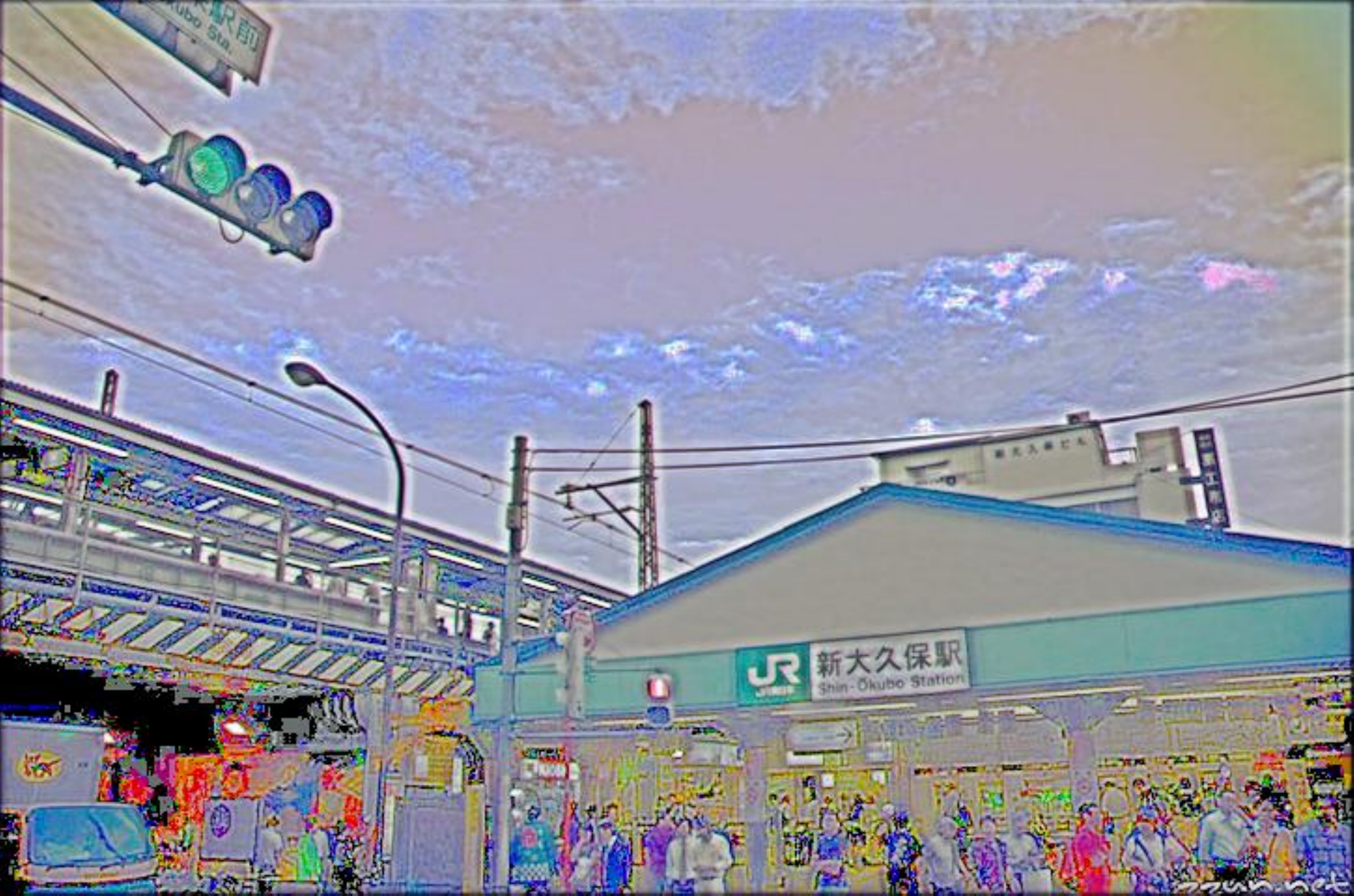}\\
			(a) Input~~~~~& (b) Zero-DCE~~~~~& (c) w/o $L_{spa}$~~~~~& (d) w/o $L_{exp}$~~~~~& (e) w/o $L_{col}$~~~~~& (f) w/o $L_{tv_\mathcal{A}}$\\
		\end{tabular}
	\end{center}
	\vspace{-0.5cm}
	\caption{Ablation study of the contribution of each loss (spatial consistency loss $L_{spa}$, exposure control loss $L_{exp}$, color constancy loss $L_{col}$, illumination smoothness loss $L_{tv_\mathcal{A}}$).}
	\label{fig:loss}
	\vspace{-0.4cm}
\end{figure*}

\vspace{-5pt}
\subsection{Non-Reference Loss Functions}

To enable zero-reference learning in DCE-Net, we propose a set of differentiable non-reference losses that allow us to evaluate the quality of enhanced images.
The following four types of losses are adopted to train our DCE-Net.

\noindent
\textbf{Spatial Consistency Loss.}
The spatial consistency loss $L_{spa}$ encourages spatial coherence of the enhanced image through preserving the difference of neighboring regions between the input image and its enhanced version:
\begin{equation}
\label{equ_4}
L_{spa}=\frac{1}{K}\sum\limits_{i=1}^K\sum\limits_{j\in\Omega(i)}(|(Y_{i}-Y_{j})|-|(I_{i}-I_{j})|)^2,
\end{equation}
where $K$ is the number of local region, and $\Omega$($i$) is the four neighboring regions (top, down, left, right) centered at the region $i$. We denote $Y$ and $I$ as the average intensity value of the local region in the enhanced version and input image, respectively.
We empirically set the size of the local region to 4$\times$4. This loss is stable given other region sizes.

\noindent
\textbf{Exposure Control Loss.}
To restrain under-/over-exposed regions, we design an exposure control loss $L_{exp}$ to control the exposure level.
The exposure control loss measures the distance between the average intensity value of a local region to the well-exposedness level $E$.
We follow existing practices~\cite{Exposure2007,Exposure2009} to set $E$ as the gray level in the RGB color space. We set $E$ to 0.6 in our experiments although we do not find much performance difference by setting $E$ within $[0.4,0.7]$.
The loss $L_{exp}$ can be expressed as:
\begin{equation}
\label{equ_5}
L_{exp}=\frac{1}{M}\sum\nolimits_{k=1}^M|Y_{k}-E|,
\end{equation}
where $M$ represents the number of nonoverlapping local regions of size 16$\times$16, $Y$ is the average intensity value of a local region in the enhanced image.

\noindent
\textbf{Color Constancy Loss.}
Following Gray-World color constancy hypothesis~\cite{Buchsbaum1980} that color in each sensor channel averages to gray over the entire image, we design a color constancy loss to correct the potential color deviations in the enhanced image and also build the relations among the three adjusted channels.
The color constancy loss $L_{col}$ can be expressed as:
\begin{equation}
\label{equ_6}
L_{col}=\sum\nolimits_{\forall(p,q)\in\varepsilon}(J^{p}-J^{q})^2, \varepsilon=\{(R,G),(R,B),(G,B)\},
\end{equation}
where $J^{p}$ denotes the average intensity value of $p$ channel in the enhanced image, ($p$,$q$) represents a pair of channels.

\noindent
\textbf{Illumination Smoothness Loss.} To preserve the monotonicity relations between neighboring pixels, we add an illumination smoothness loss to each curve parameter map $\mathcal{A}$. The illumination smoothness loss $L_{tv_\mathcal{A}}$ is defined as:
\begin{equation}
\label{equ_7}
L_{tv_\mathcal{A}}=\frac{1}{N}\sum\limits_{n=1}^N\sum\limits_{c\in\xi}(|\nabla_{x}\mathcal{A}_{n}^{c}|+\nabla_{y}\mathcal{A}_{n}^{c}|)^2,  \xi=\{R,G,B\},
\end{equation}
where $N$ is the number of iteration, $\nabla_{x}$ and $\nabla_{y}$ represent the horizontal and vertical gradient operations, respectively.


\noindent
\textbf{Total Loss.}
The total loss can be expressed as:
\begin{equation}
\label{equ_9}
L_{total}=L_{spa}+L_{exp}+W_{col}L_{col}+W_{tv_\mathcal{A}}L_{tv_\mathcal{A}},
\end{equation}
where $W_{col}$ and $W_{tv_\mathcal{A}}$ are the weights of the losses.

\section{Experiments}

\noindent
\textbf{Implementation Details.}
CNN-based models usually use self-captured paired data for network training \cite{Chen2018,Chenchen2018,Wang2019,LightenNet,wang2019salient,xu2020deep} while GAN-based models elaborately select unpaired data \cite{Jiang2019,lcy2018spl,Yu2018,Chencvpr2018,Ignatov2018}. To bring the capability of wide dynamic range adjustment into full play, we incorporate both low-light and over-exposed images into our training set. To this end, we employ 360 multi-exposure sequences from the Part1 of SICE dataset~\cite{Cai2018} to train the proposed DCE-Net. The dataset is also used as a part of the training data in EnlightenGAN \cite{Jiang2019}.
%
%
We randomly split 3,022 images of different exposure levels in the Part1 subset~\cite{Cai2018} into two parts (2,422 images for training and the rest for validation).
We resize the training images to the size of 512$\times$512.
%

We implement our framework with PyTorch on an NVIDIA $2080$Ti GPU.
A batch size of $8$ is applied.
The filter weights of each layer are initialized with standard zero mean and 0.02 standard deviation Gaussian function.
Bias is initialized as a constant.
We use ADAM optimizer with default parameters and fixed learning rate $1e^{-4}$ for our network optimization.
The weights $W_{col}$ and $W_{tv_\mathcal{A}}$ are set to 0.5, and 20, respectively, to balance the scale of losses.

%
%

\vspace{-5pt}
\subsection{Ablation Study}

We perform several ablation studies to demonstrate the effectiveness of each component of Zero-DCE as follows.
More qualitative and quantitative comparisons can be found in the supplementary material.


\noindent
\textbf{Contribution of Each Loss.}
We present the results of Zero-DCE trained by various combinations of losses in Fig.~\ref{fig:loss}.
The result without spatial consistency loss $L_{spa}$ has relatively lower contrast (\eg, the cloud regions) than the full result. This shows the importance of $L_{spa}$ in preserving the difference of neighboring regions between the input and the enhanced image.
Removing the exposure control loss $L_{exp}$ fails to recover the low-light region.
Severe color casts emerge when the color constancy loss $L_{col}$ is discarded. This variant ignores the relations among three channels when curve mapping is applied.
Finally, removing the illumination smoothness loss $L_{tv_\mathcal{A}}$ hampers the correlations between neighboring regions leading to obvious artifacts.

\begin{figure}
	\begin{center}
		\begin{tabular}{c@{ }c@{ }c@{ }c@{ }c@{ }c@{ }c}
			\includegraphics[width=.15\textwidth,height=2cm]{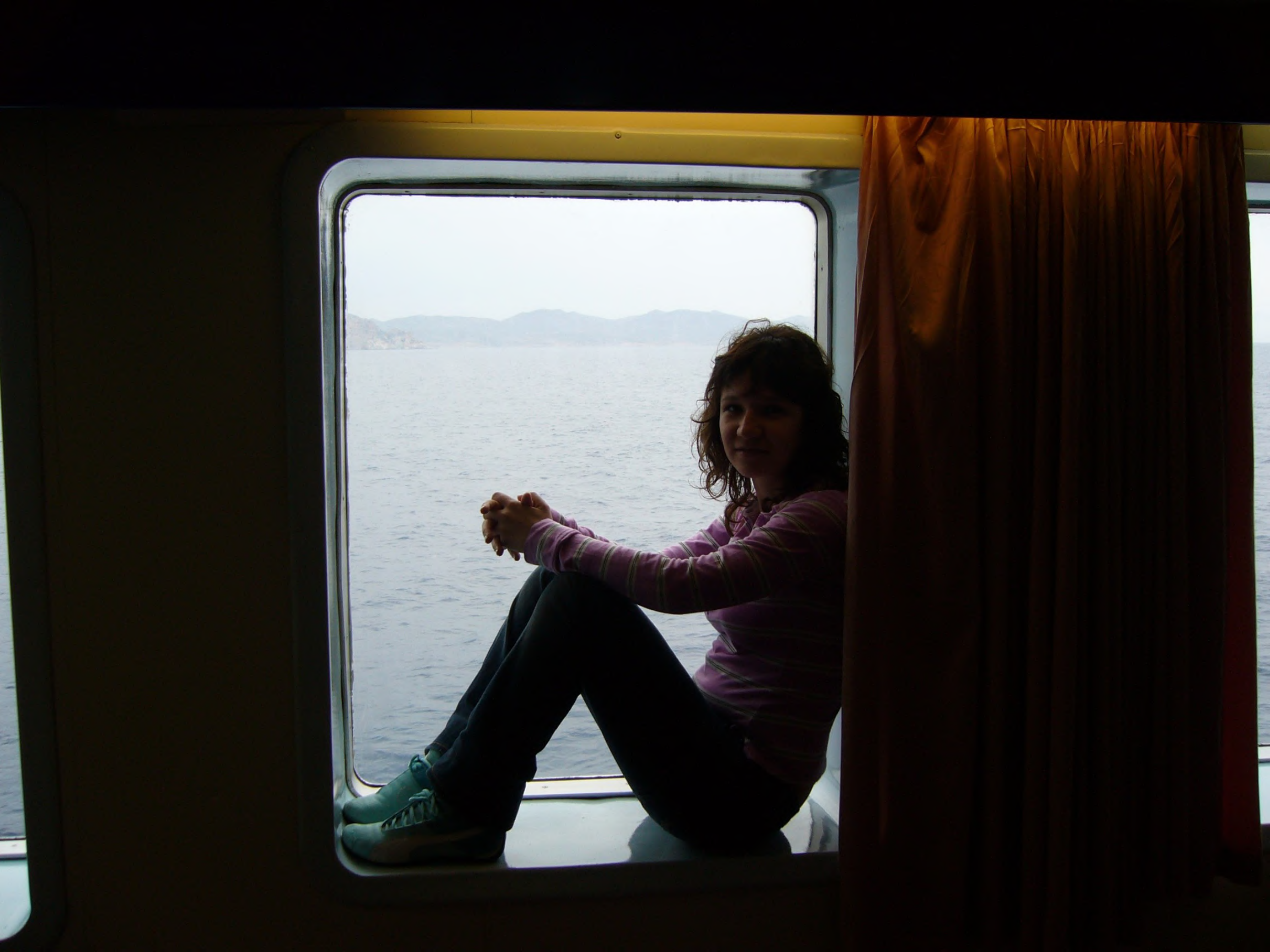}~&
			\includegraphics[width=.15\textwidth,height=2cm]{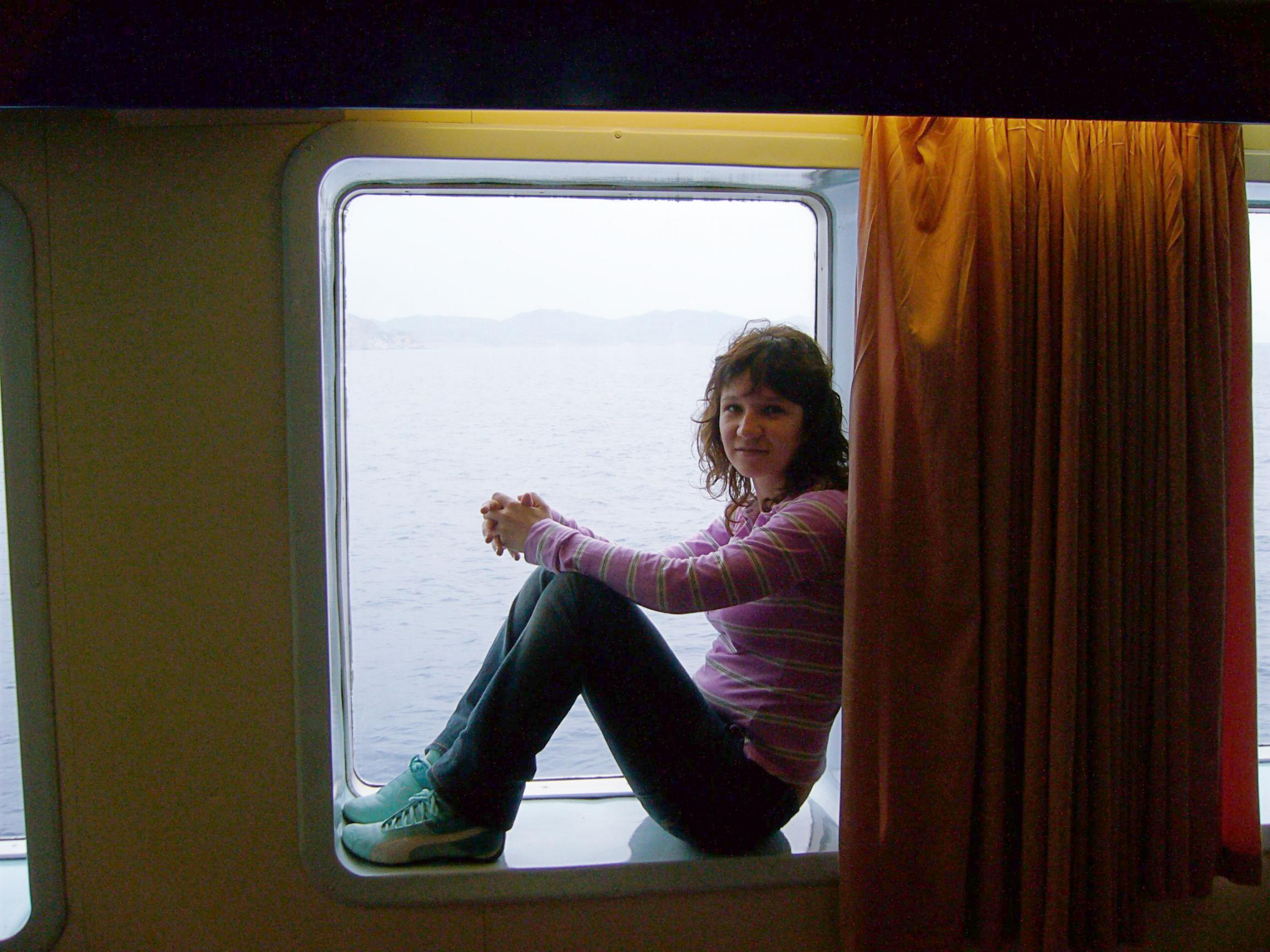}~&
			\includegraphics[width=.15\textwidth,height=2cm]{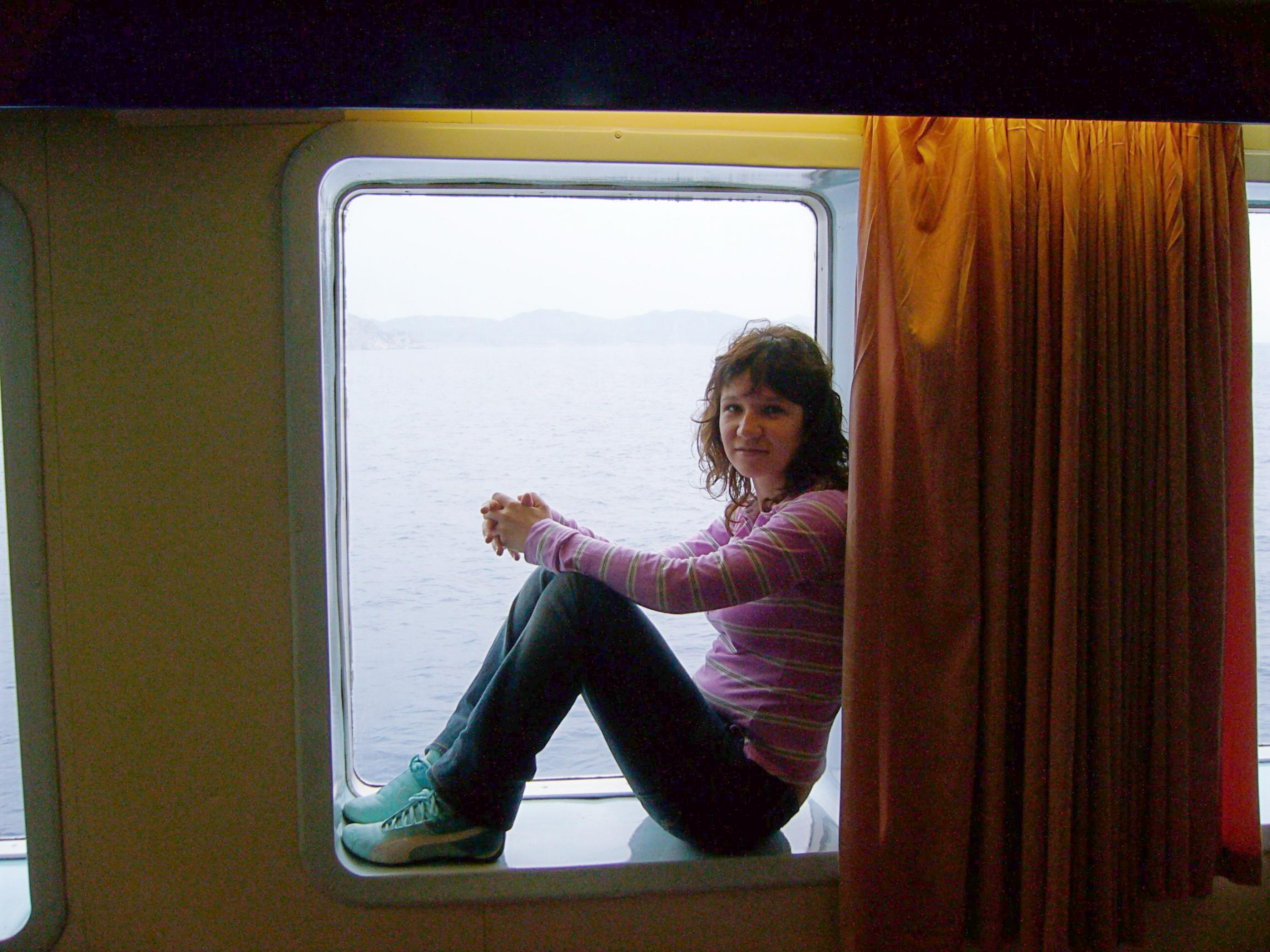}\\
			(a) Input~& (b) 3-32-8~& (c) 7-16-8\\
			\includegraphics[width=.15\textwidth,height=2cm]{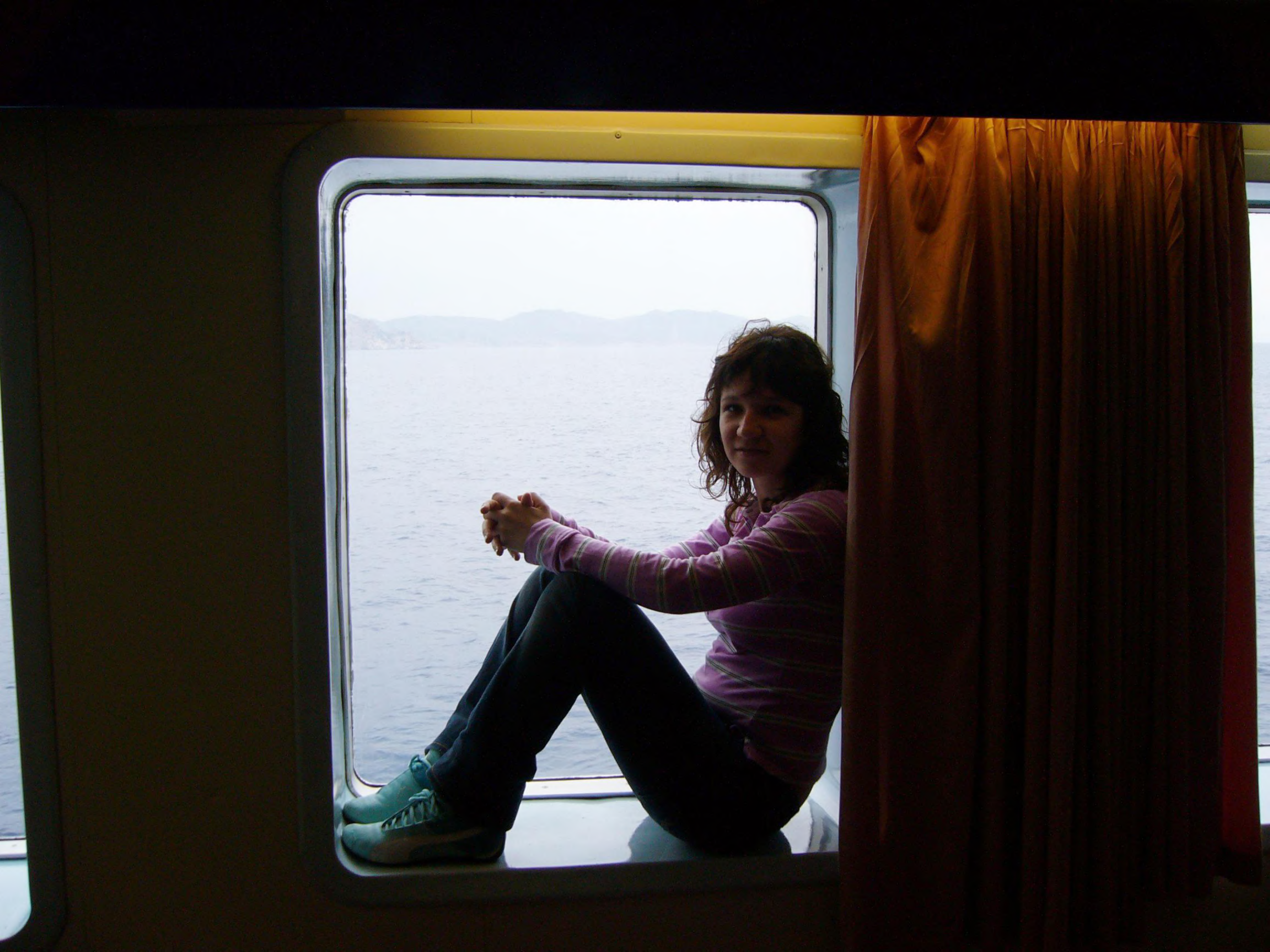}~&
			\includegraphics[width=.15\textwidth,height=2cm]{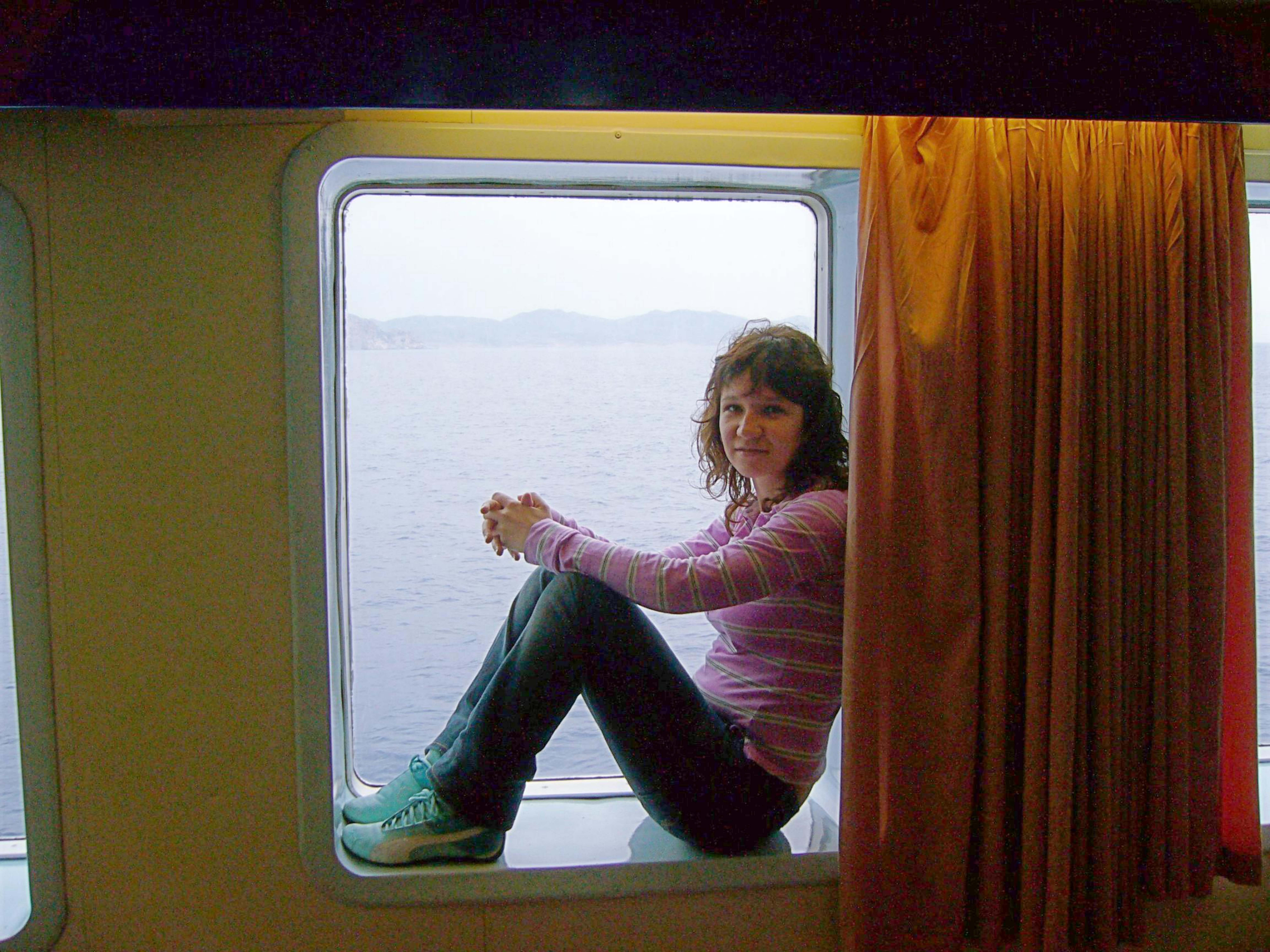}~&
			\includegraphics[width=.15\textwidth,height=2cm]{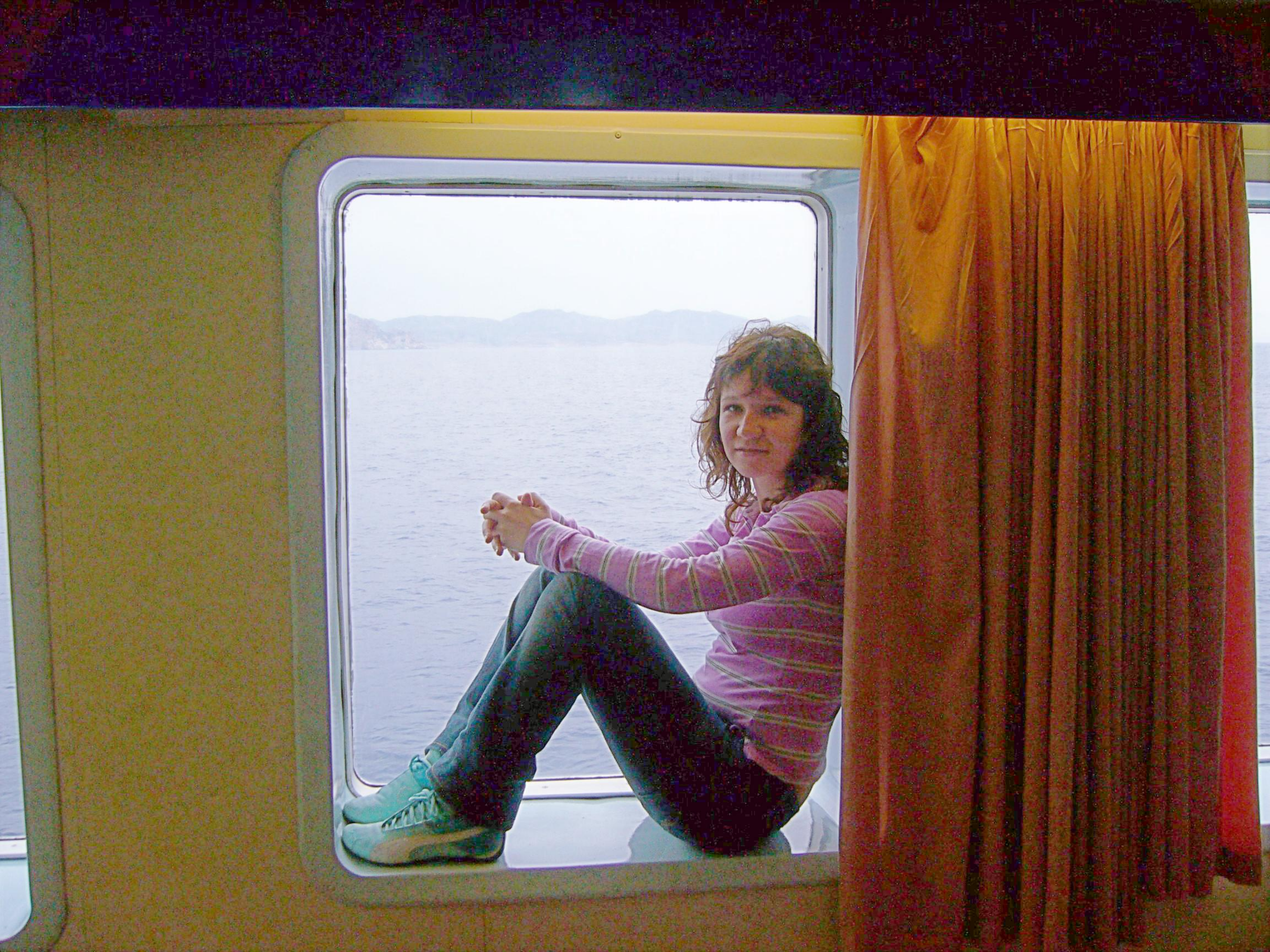}\\
			(d) 7-32-1~& (e) 7-32-8 & (f) 7-32-16\\
		\end{tabular}
	\end{center}
	\vspace{-0.5cm}
	\caption{Ablation study of the effect of parameter settings. $l$-$f$-$n$ represents the proposed Zero-DCE with $l$ convolutional layers, $f$ feature maps of each layer (except the last layer), and $n$ iterations.}
	\label{fig:parameter}
	\vspace{-0.7cm}
\end{figure}

\noindent
\textbf{Effect of Parameter Settings.} We evaluate the effect of parameters in Zero-DCE, consisting of the depth and width of the DCE-Net and the number of iterations.
A visual example is presented in Fig.~\ref{fig:parameter}.
In Fig.~\ref{fig:parameter}(b), with just three convolutional layers, Zero-DCE$_{3-32-8}$ can already produce satisfactory results, suggesting the effectiveness of zero-reference learning.
The Zero-DCE$_{7-32-8}$ and Zero-DCE$_{7-32-16}$ produce most visually pleasing results with natural exposure and proper contrast.
By reducing the number of iterations to 1, an obvious decrease in performance is observed on Zero-DCE$_{7-32-1}$ as shown in Fig.~\ref{fig:parameter}(d).
This is because the curve with only single iteration has limited adjustment capability. This suggests the need for higher-order curves in our method.
We choose Zero-DCE$_{7-32-8}$ as the final model based given its good trade-off between efficiency and restoration performance.
%

\begin{figure*}[t]
	\begin{center}
		\begin{tabular}{c@{ }c@{ }c@{ }c@{ }c@{ }c}
			\includegraphics[width=.18\textwidth,height=2.3cm]{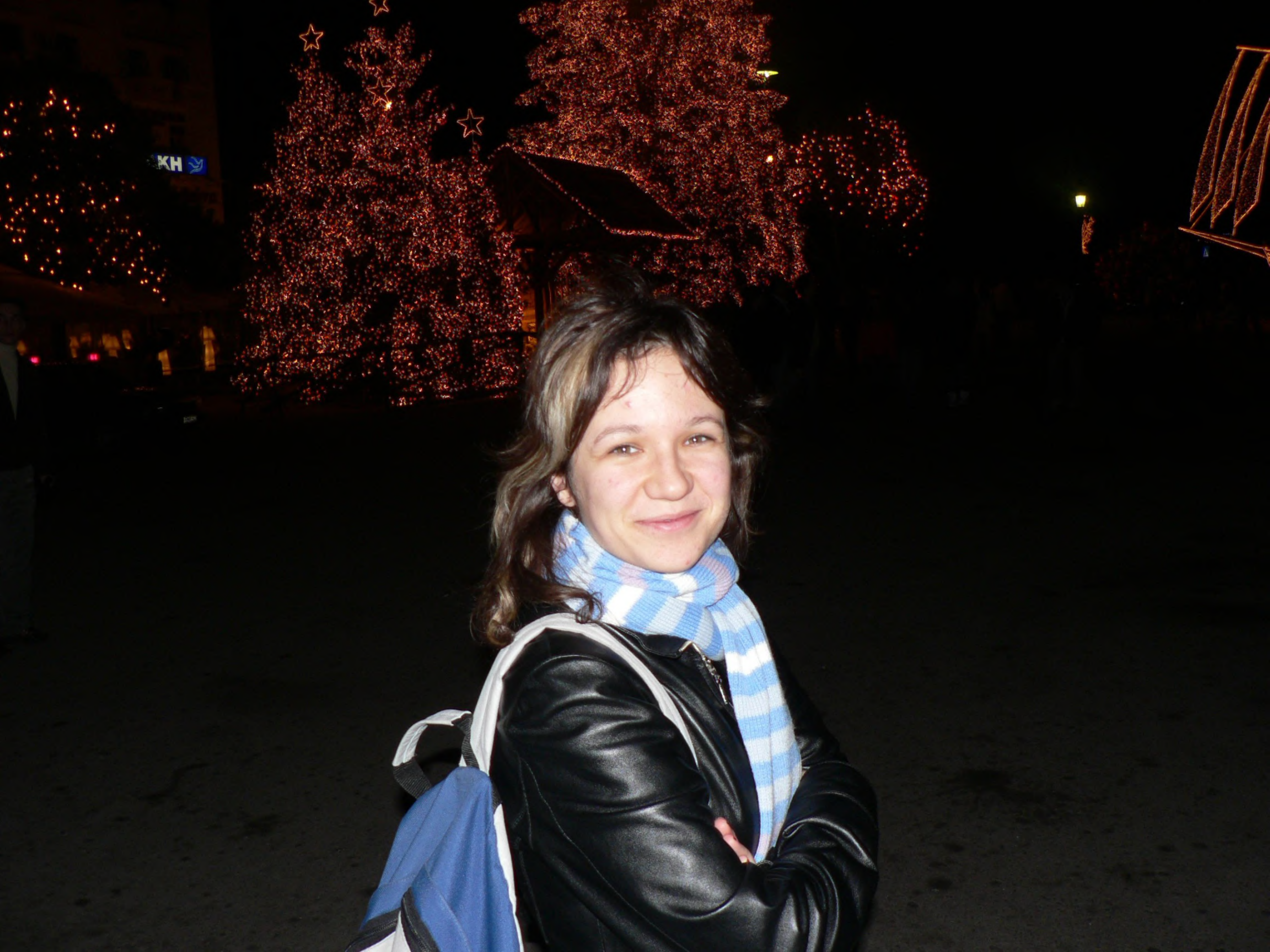}~&
			\includegraphics[width=.18\textwidth,height=2.3cm]{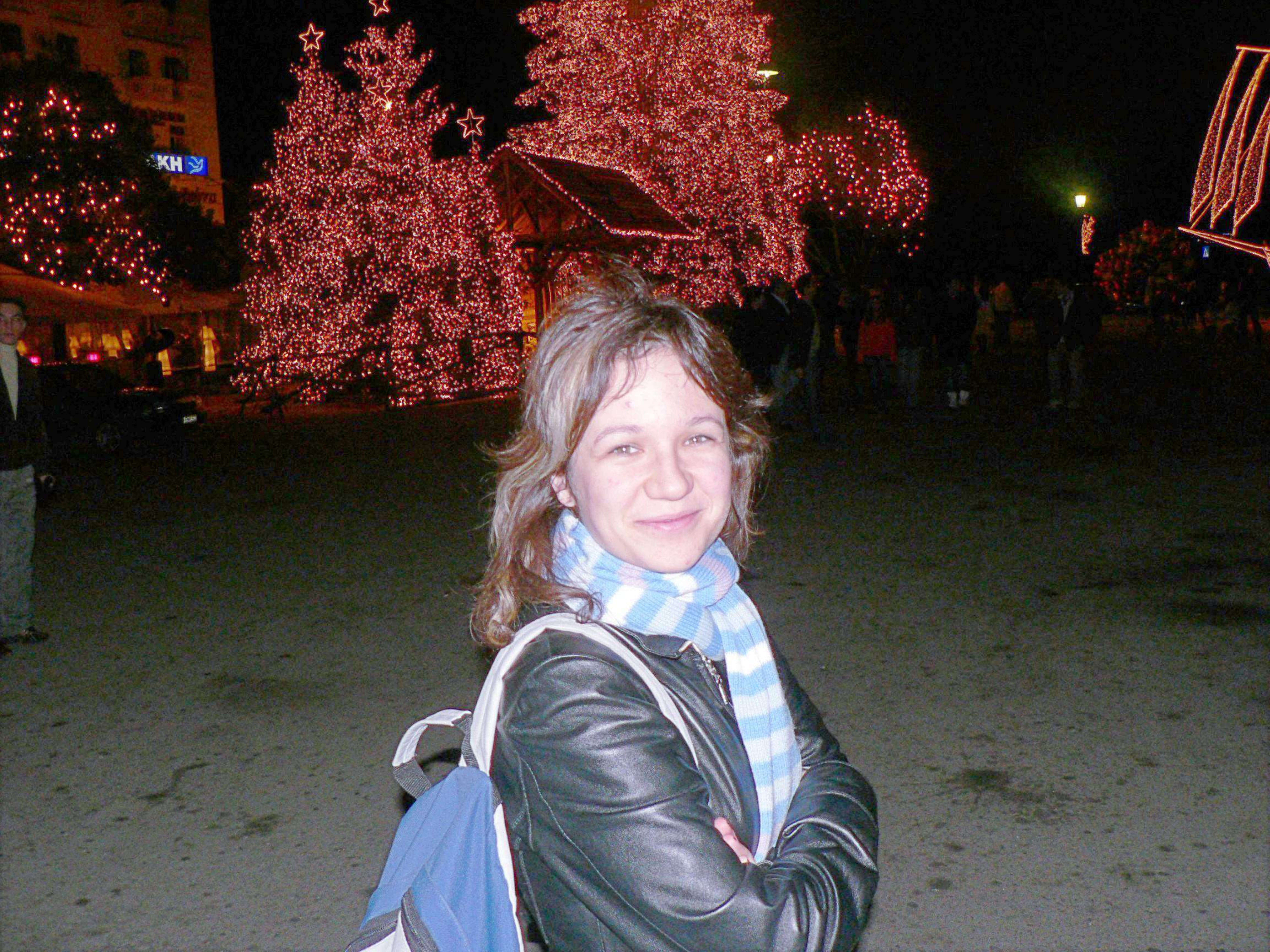}~&
			\includegraphics[width=.18\textwidth,height=2.3cm]{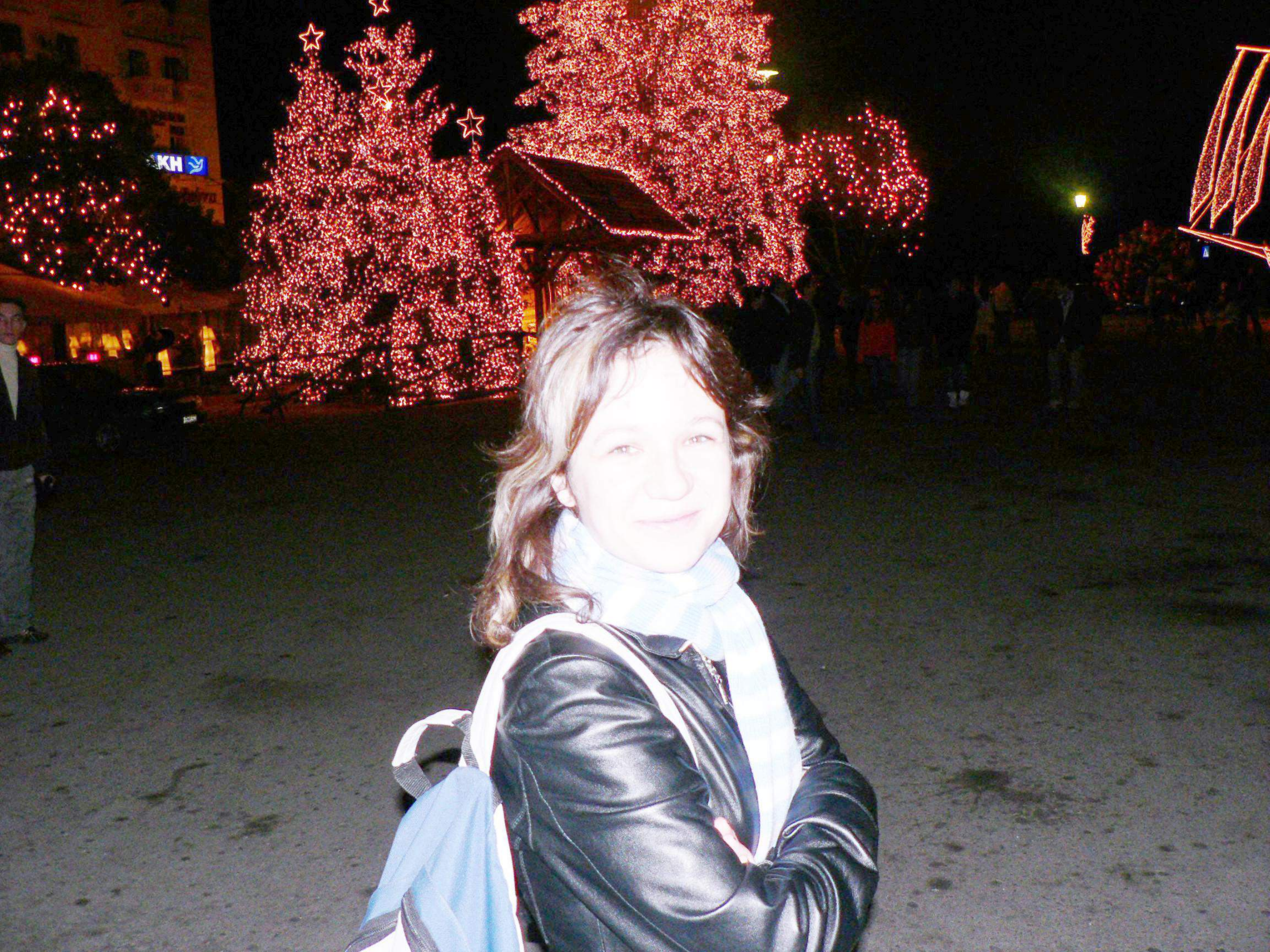}~&
			\includegraphics[width=.18\textwidth,height=2.3cm]{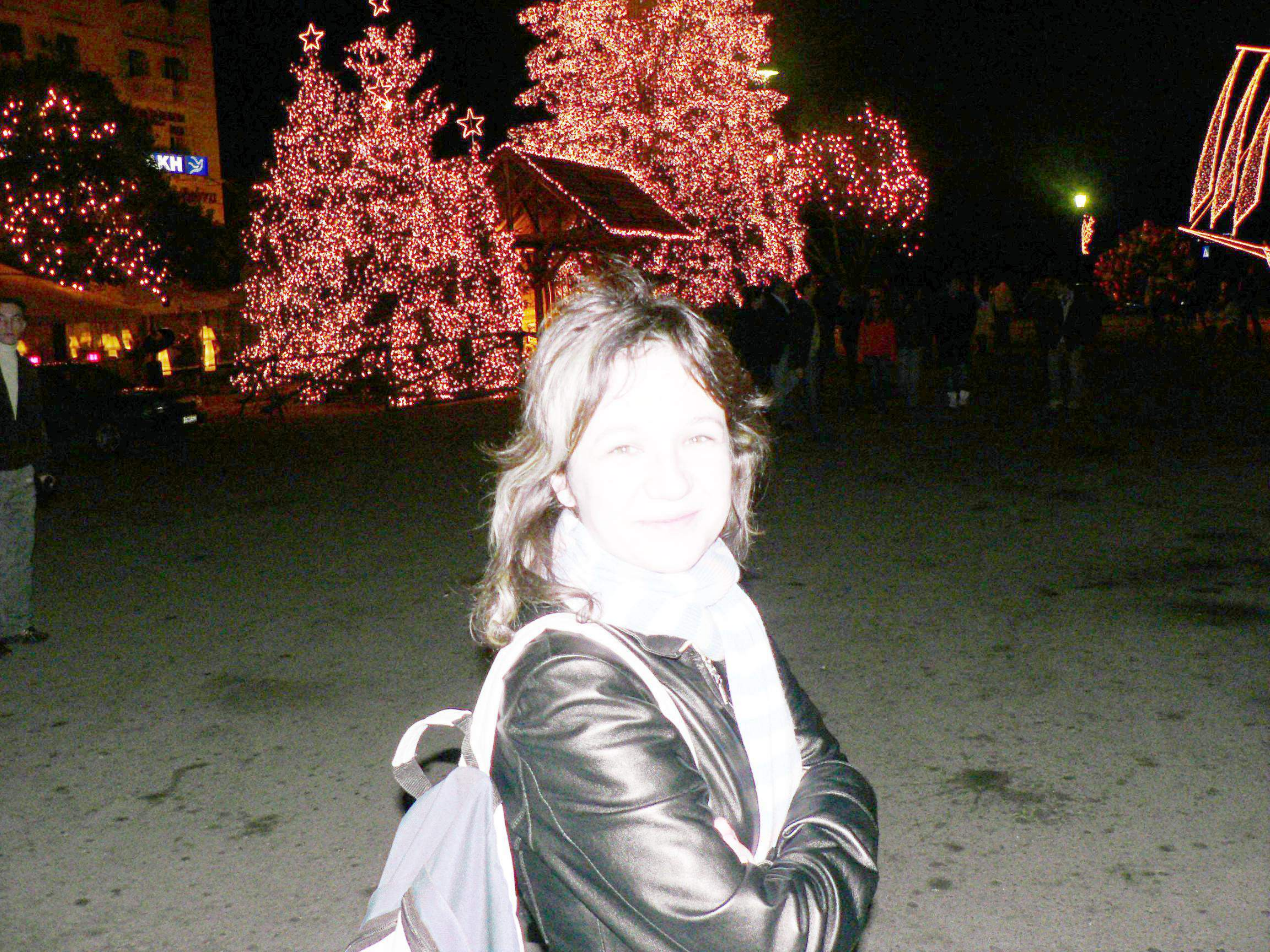}~&
			\includegraphics[width=.18\textwidth,height=2.3cm]{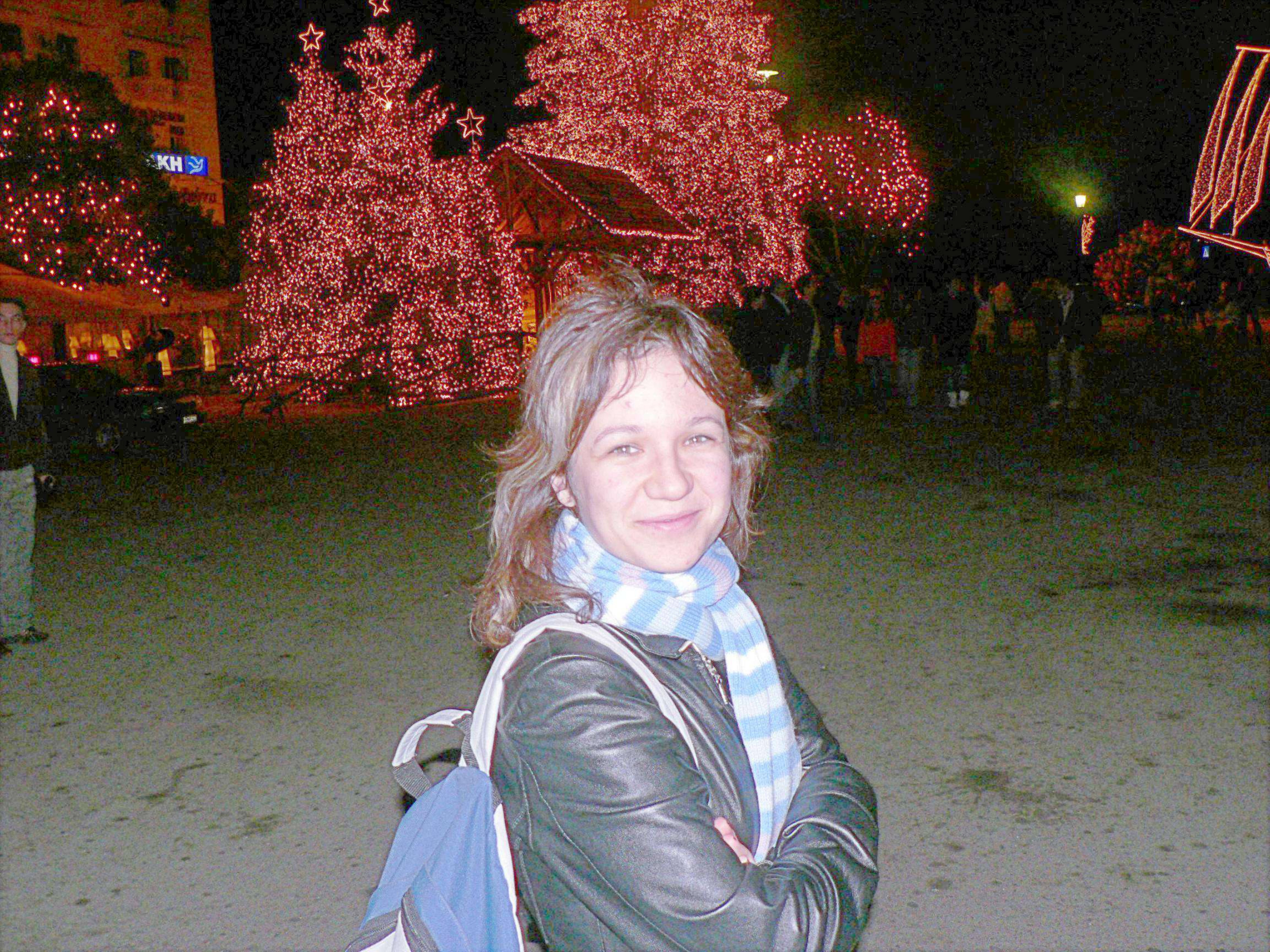}\\
			(a) Input~& (b) Zero-DCE~& (c) Zero-DCE$_{Low}$~& (d) Zero-DCE$_{LargeL}$~& (e) Zero-DCE$_{LargeLH}$\\
		\end{tabular}
	\end{center}
	\vspace{-0.5cm}
	\caption{Ablation study on the impact of training data.}
	\vspace{-0.2cm}
	\label{fig:trainingdata}
\end{figure*}

\noindent
\textbf{Impact of Training Data.}
To test the impact of training data, we retrain the Zero-DCE on  different datasets: 1) only 900 low-light images out of 2,422 images in the original training set (Zero-DCE$_{Low}$), 2) 9,000 unlabeled low-light images provided in the DARK FACE dataset~\cite{2019arXiv190404474Y} (Zero-DCE$_{LargeL}$), and 3) 4800 multi-exposure images from the data augmented combination of Part1 and Part2 subsets in the SICE dataset~\cite{Cai2018} (Zero-DCE$_{LargeLH}$).
%
%
%
As shown in Fig.~\ref{fig:trainingdata}(c) and (d), after removing the over-exposed training data, Zero-DCE tends to over-enhance the well-lit regions (\eg, the face), in spite of using more low-light images, (\ie, Zero-DCE$_{LargeL}$). Such results indicate the rationality and necessity of the usage of multi-exposure training data in the training process of our network.  In addition, the Zero-DCE can better recover the dark regions when more multi-exposure training data are used (\ie, Zero-DCE$_{LargeLH}$), as shown in Fig.~\ref{fig:trainingdata}(e). For a fair comparison with other deep learning-based methods, we use a comparable amount of training data with them although more training data can bring better visual performance to our approach.

\begin{figure*}
	\begin{center}
		\begin{tabular}{c@{ }c@{ }c@{ }c@{ }c@{ }}
			\includegraphics[height=4.5cm,width=3.5cm]{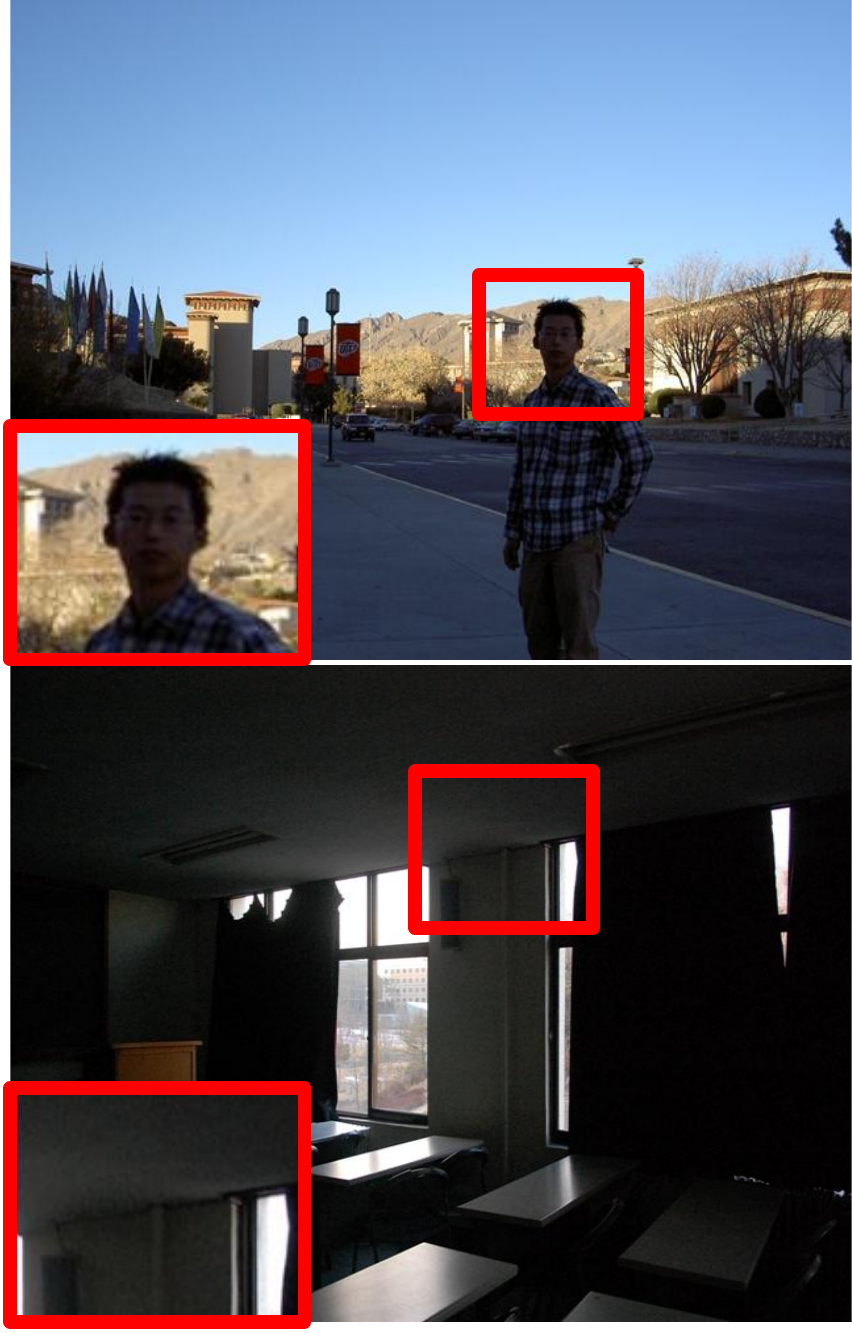}&~~
			\includegraphics[height=4.5cm,width=3.5cm]{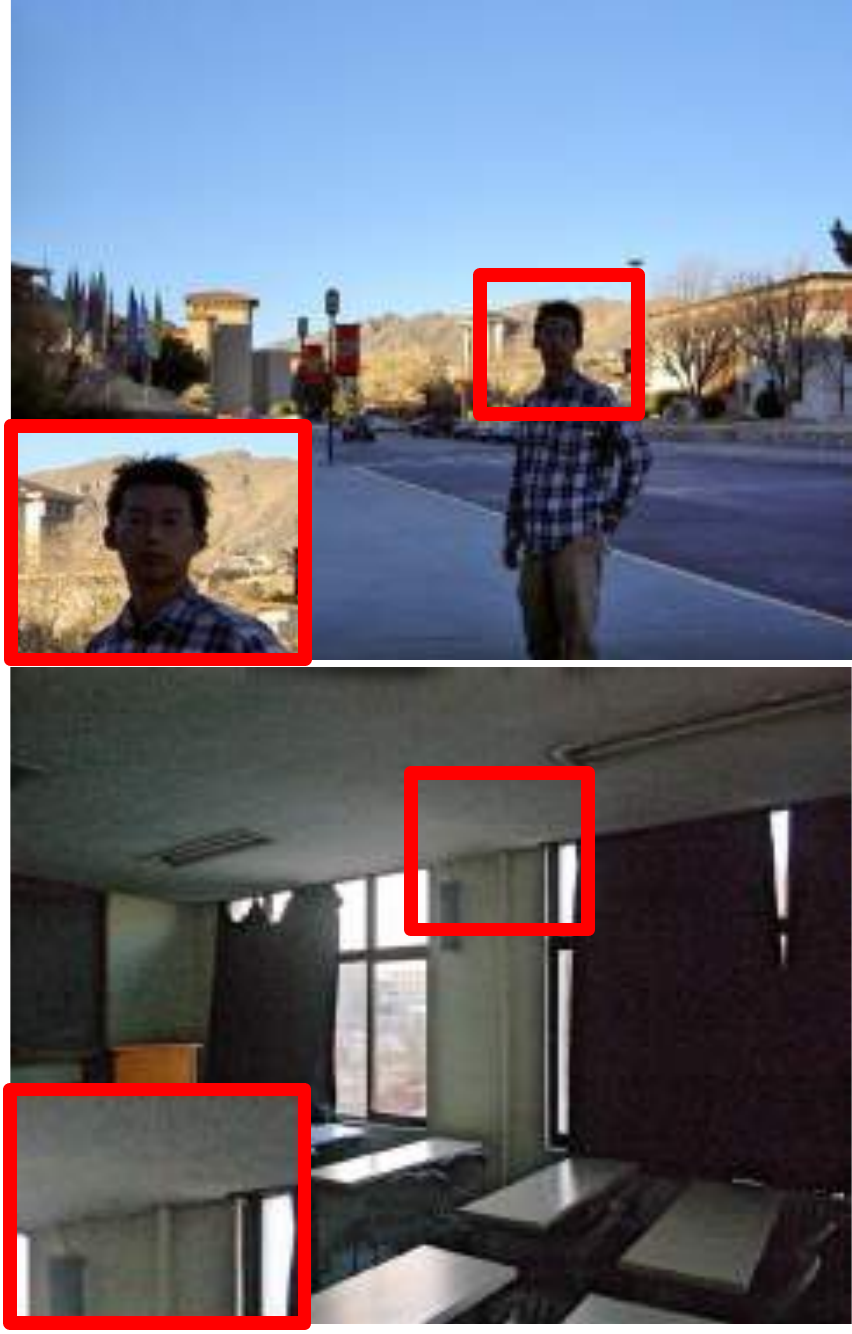}&~~
			\includegraphics[height=4.5cm,width=3.5cm]{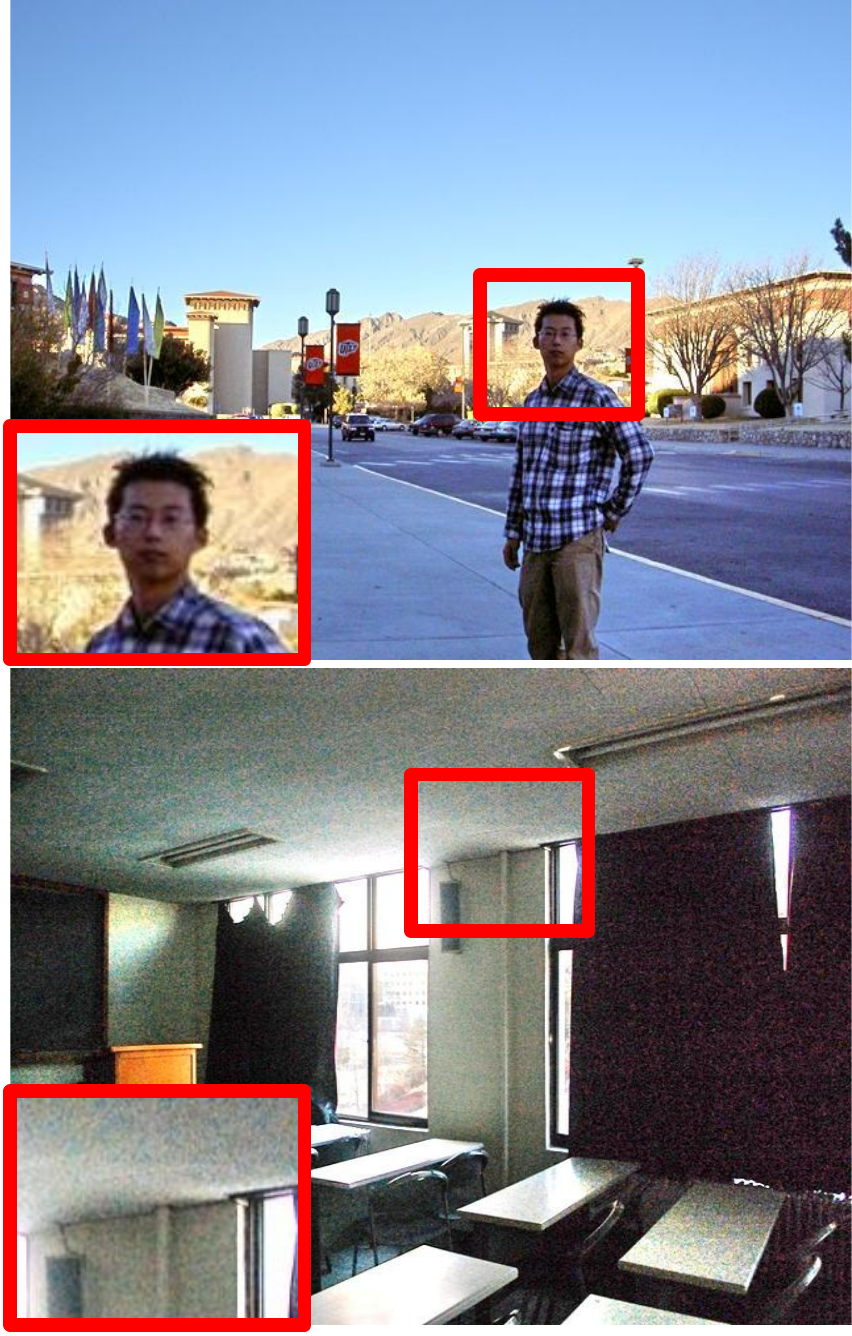}&~~
			\includegraphics[height=4.5cm,width=3.5cm]{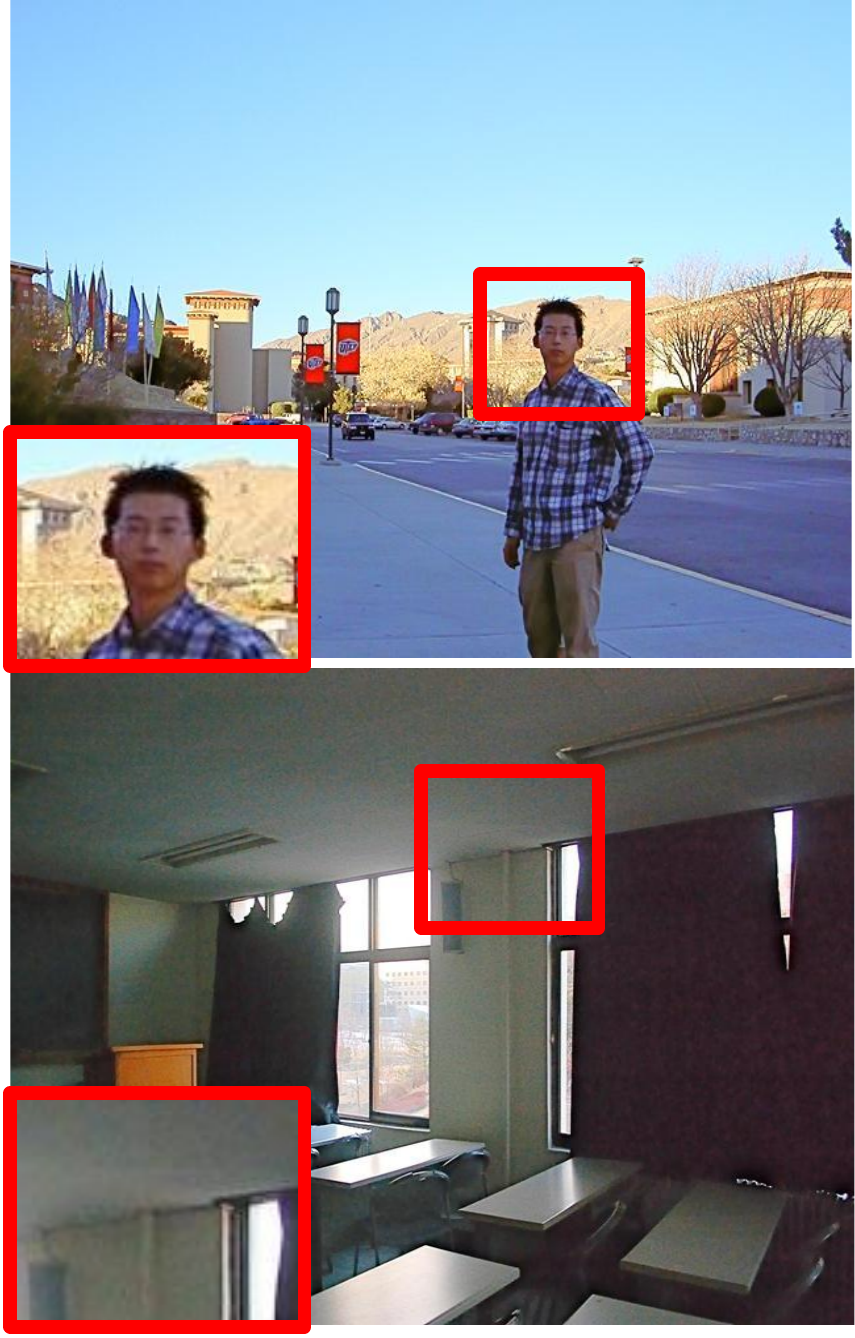}\\
			(a) Inputs &~~  (b) SRIE~\cite{Fu2016}  &~~ (c)  LIME~\cite{Guo2017} &~~ (d) Li \etal~\cite{Li2018} \\
			\includegraphics[height=4.5cm,width=3.5cm]{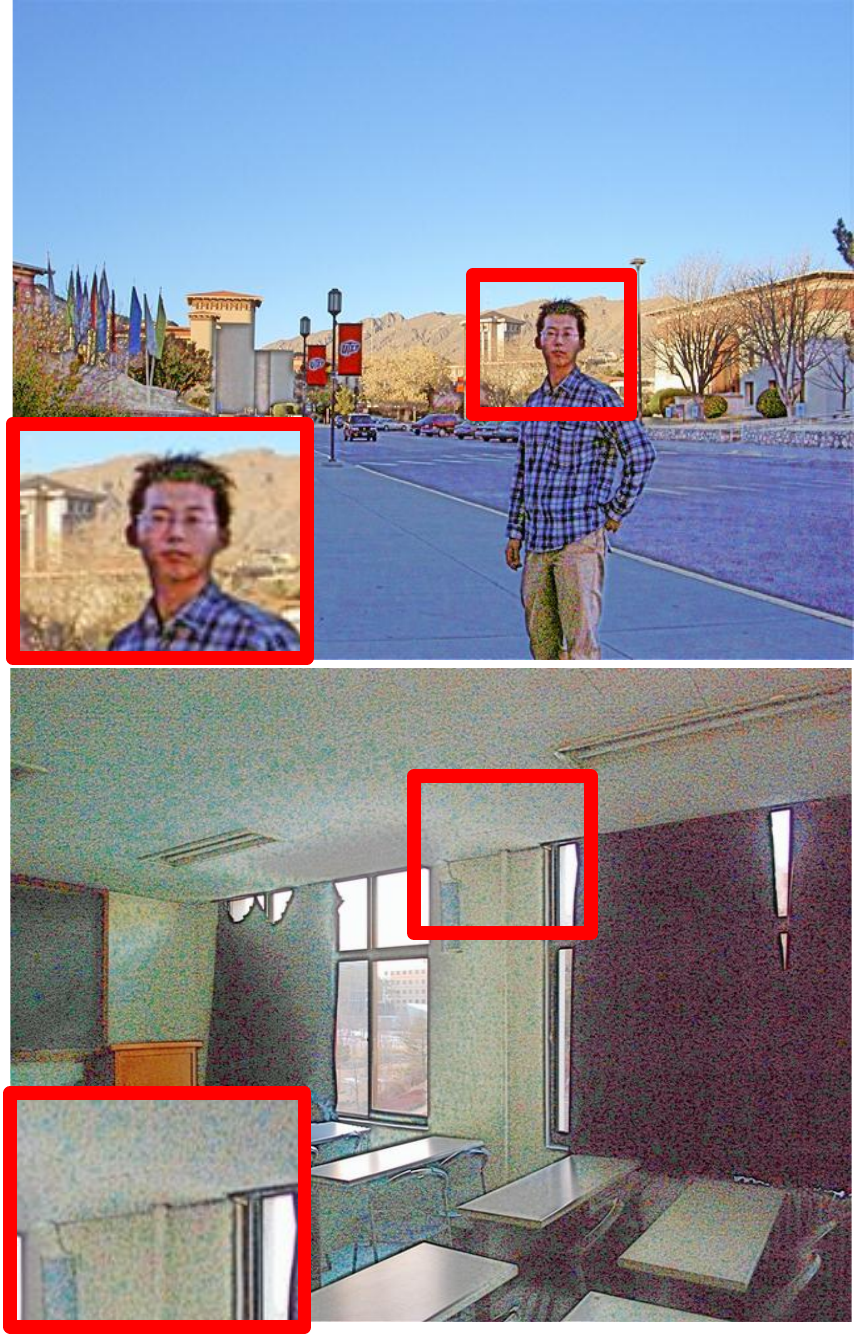}&~~
			\includegraphics[height=4.5cm,width=3.5cm]{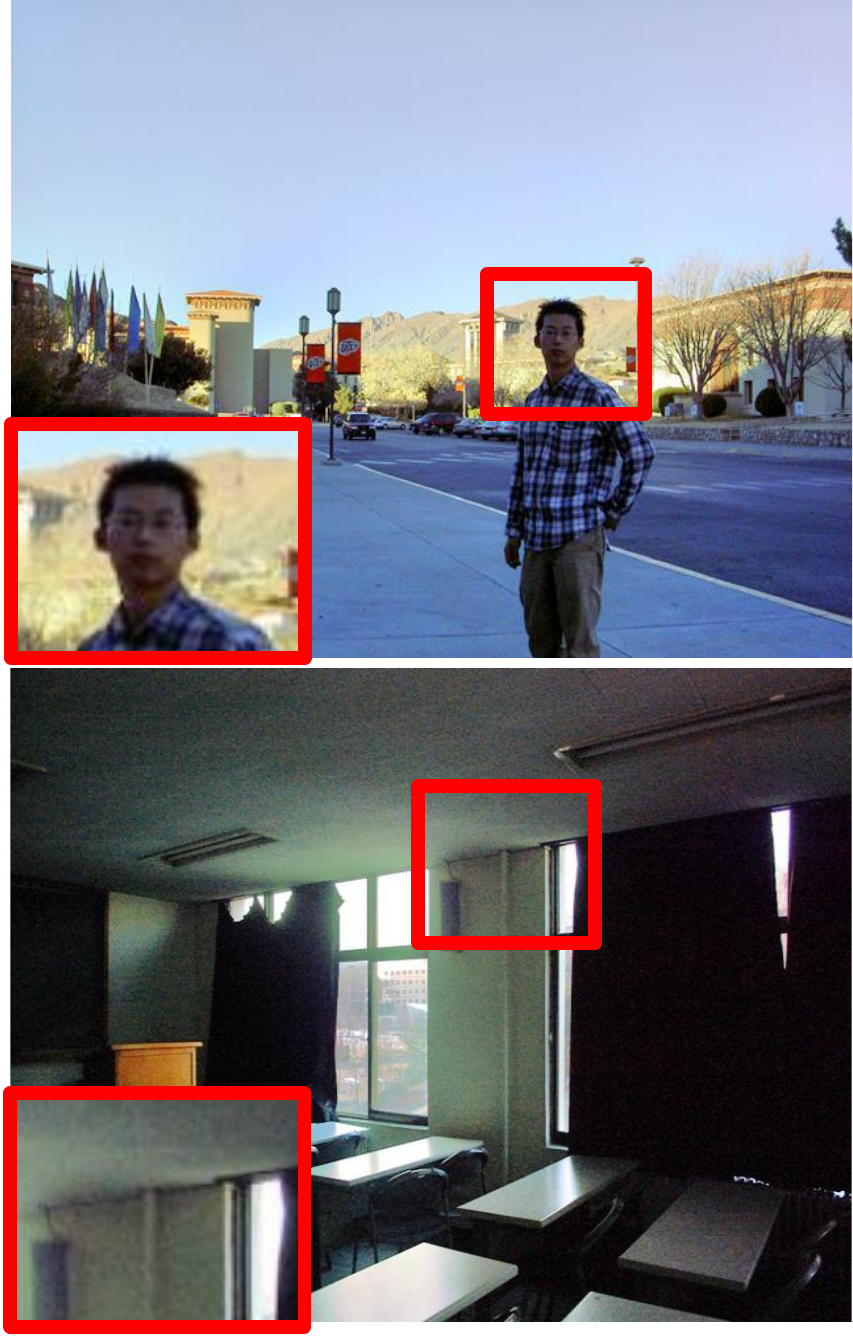}&~~
			\includegraphics[height=4.5cm,width=3.5cm]{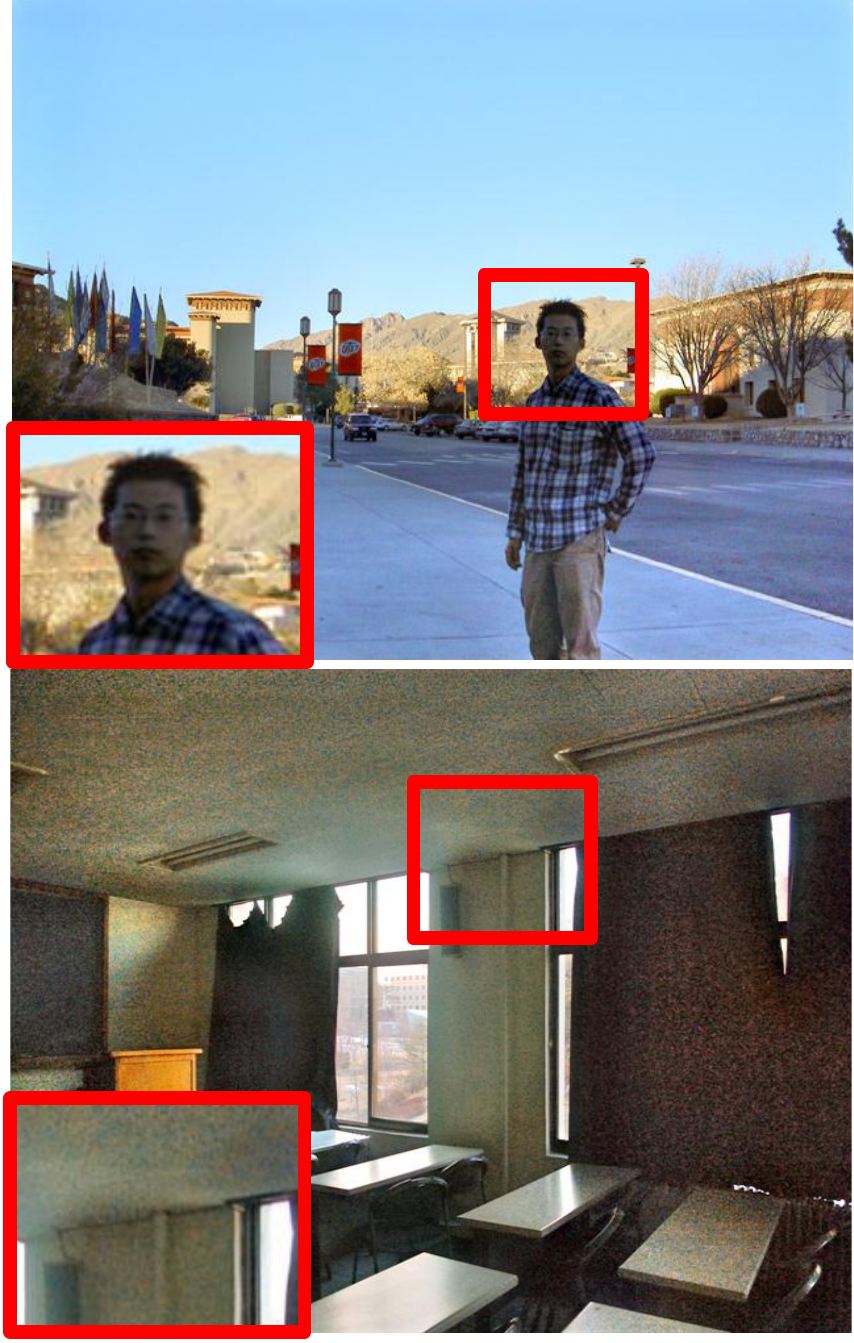}&~~
			\includegraphics[height=4.5cm,width=3.5cm]{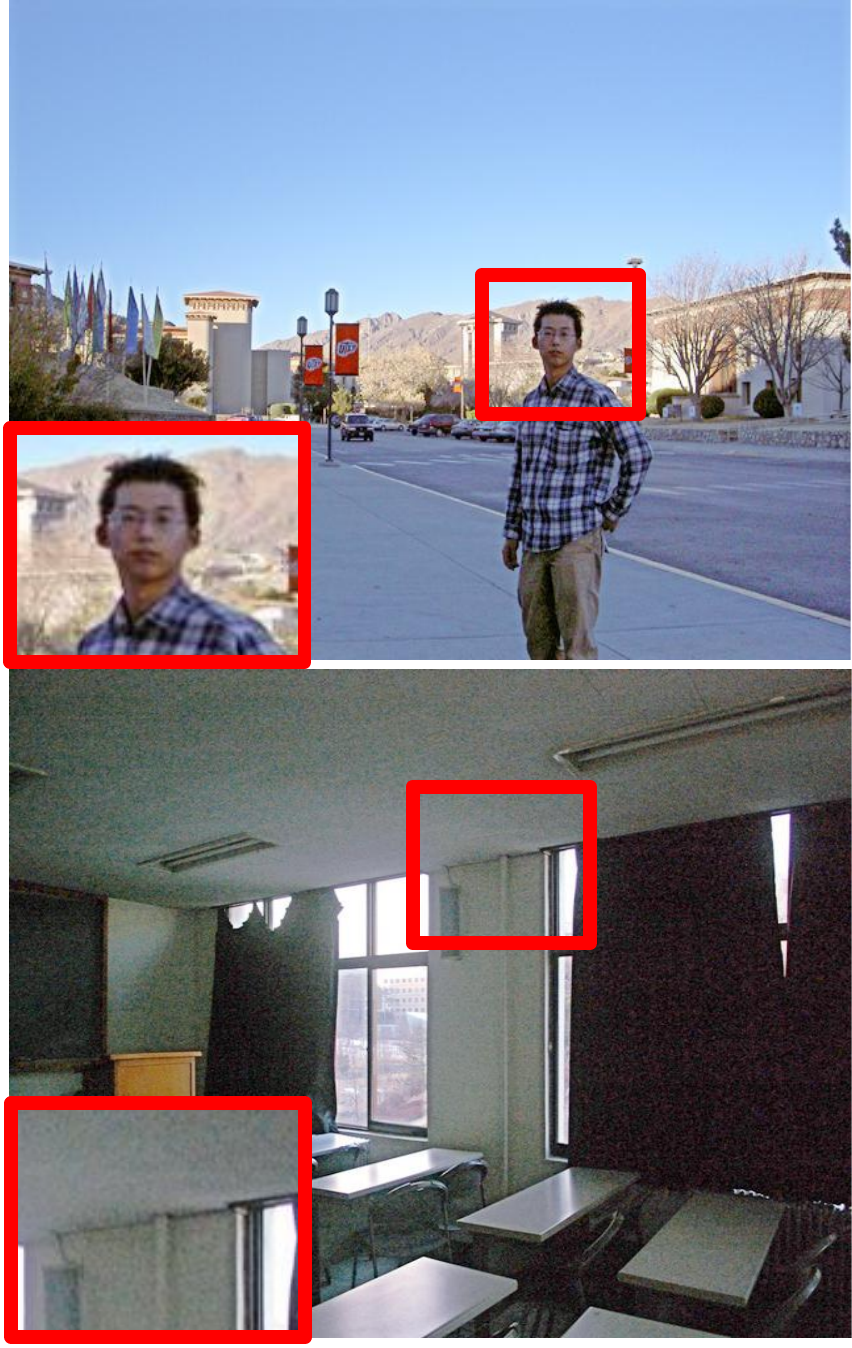}\\
			(e) RetinexNet~\cite{Chen2018} &~~ (f)  Wang \etal~\cite{Wang2019} &~~ (g) EnlightenGAN~\cite{Jiang2019} &~~ (h) Zero-DCE\\
		\end{tabular}
	\end{center}
	\vspace{-0.5cm}
	\caption{Visual comparisons on typical low-light images. Red boxes indicate the obvious differences. }
	\label{fig:visual_results}
	\vspace{-0.5cm}
\end{figure*}

\subsection{Benchmark Evaluations}
We compare Zero-DCE with several state-of-the-art methods: three conventional methods (SRIE~\cite{Fu2016}, LIME~\cite{Guo2017}, Li \etal~\cite{Li2018}), two CNN-based methods (RetinexNet~\cite{Chen2018}, Wang \etal~\cite{Wang2019} ), and one GAN-based method (EnlightenGAN~\cite{Jiang2019}).
The results are reproduced by using publicly available source codes with recommended parameters.

We perform qualitative and quantitative experiments on standard image sets from previous works including NPE~\cite{Wang2013} (84 images), LIME~\cite{Guo2017} (10 images), MEF~\cite{Ma2015} (17 images), DICM~\cite{Lee2012} (64 images), and VV\footnote[3]{\url{https://sites.google.com/site/vonikakis/datasets}}  (24 images).
%
%
Besides, we quantitatively validate our method on the Part2 subset of SICE dataset~\cite{Cai2018}, which consists of 229 multi-exposure sequences and the corresponding reference image for each multi-exposure sequence.
For a fair comparison, we only use the low-light images of Part2 subset~\cite{Cai2018} for testing, since baseline methods cannot handle over-exposed images well.
Specifically, we choose the first three (resp. four) low-light images if there are seven (resp. nine) images in a multi-exposure sequence and resize all images to a size of 1200$\times$900$\times$3.
Finally, we obtain 767 paired low/normal light images.
We discard the low/normal light image dataset mentioned in~\cite{2019arXiv190404474Y}, because the training datasets of RetinexNet~\cite{Chen2018} and EnlightenGAN~\cite{Jiang2019} consist of some images from this dataset. Note that the latest paired training and testing dataset constructed in \cite{Wang2019} are not publicly available. We did not use the MIT-Adobe FiveK dataset \cite{Adobe5K} as it is not primarily designed for underexposed photos enhancement.
%
%
%
%

\begin{table*}[htbp]
	\caption{User study (US)$\uparrow$/Perceptual index (PI)$\downarrow$ scores on the image sets (NPE, LIME, MEF, DICM, VV). Higher US score indicates better human subjective visual quality while lower PI value indicates better perceptual quality. The best result is in red whereas the second best one is in blue under each case.}
	\vspace{-0.2cm}
	\centering
	\begin{tabular}{c|c|c|c|c|c|c}
		\hline
		\textbf{Method}               & \textbf{NPE} & \textbf{LIME} &\textbf{MEF}  & \textbf{DICM} & \textbf{VV} & \textbf{Average}\\
		\hline
		SRIE~\cite{Fu2016}           &  3.65/{\color{red}2.79} & 3.50/{\color{red}2.76}               &  3.22/2.61           &  3.42/3.17             & 2.80/3.37                  &3.32/{\color{blue}2.94} \\
		LIME~\cite{Guo2017}          &  3.78/3.05           & {\color{red}3.95}/3.00  &  3.71/2.78            &  3.31/3.35             & {\color{blue}3.21}/{\color{blue}3.03} & 3.59/3.04\\
		Li \etal~\cite{Li2018}           &  3.80/3.09           &3.78/3.02       & 2.93/3.61                &3.47/3.43                & 2.87/3.37                        & 3.37/3.72\\
		RetinexNet~\cite{Chen2018}   &  3.30/3.18           & 2.32/3.08         &2.80/2.86        &  2.88/3.24       & 1.96/{\color{red}2.95}              &2.58/3.06\\
		Wang \etal~\cite{Wang2019}      & {\color{blue}3.83}/{\color{blue}2.83}  & 3.82/2.90     &3.13/2.72           & 3.44/3.20      &  2.95/3.42 & 3.43/3.01  \\
		EnlightenGAN~\cite{Jiang2019}&  {\color{red}3.90}/2.96 & {\color{blue}3.84}/{\color{blue}2.83} & {\color{blue}3.75}/{\color{blue}2.45} &{\color{blue}3.50}/{\color{blue}3.13} & 3.17/4.71              &{\color{blue}3.63}/3.22\\
		Zero-DCE                     &  3.81/2.84 & 3.80/{\color{red}2.76}               &  {\color{red}4.13}/{\color{red}2.43}&  {\color{red}3.52}/{\color{red}3.04} & {\color{red}3.24}/3.33     &{\color{red}3.70}/{\color{red}2.88}\\
		\hline
	\end{tabular}
	\label{label:user study}
		\vspace{-11pt}
\end{table*}

\vspace{-5pt}
\subsubsection{Visual and Perceptual Comparisons}
We present the visual comparisons on typical low-light images in Fig.~\ref{fig:visual_results}.
For challenging back-lit regions (\eg, the face in Fig.~\ref{fig:visual_results}(a)), Zero-DCE yields natural exposure and clear details while SRIE~\cite{Fu2016}, LIME~\cite{Guo2017}, Wang \etal~\cite{Wang2019}, and EnlightenGAN~\cite{Jiang2019} cannot recover the face clearly. RetinexNet~\cite{Chen2018} produces over-exposed artifacts.
%
%
In the second example featuring an indoor scene, our method enhances dark regions and preserves color of the input image simultaneously. The result is visually pleasing without obvious noise and color casts.
In contrast,  Li \etal~\cite{Li2018} over-smoothes the details while other baselines amplify noise and even produce color deviation (\eg, the color of wall).
%
%

We perform a user study to quantify the subjective visual quality of various methods.
We process low-light images from the image sets (NPE, LIME, MEF, DICM, VV) by different methods.
For each enhanced result, we display it on a screen and provide the input image as a reference.
A total of 15 human subjects are invited to independently score the visual quality of the enhanced image.
%
These subjects are trained by observing the results from 1) whether the results contain over-/under-exposed artifacts or over-/under-enhanced regions; 2) whether the results introduce color deviation; and 3) whether the results have unnatural texture and obvious noise.
The scores of visual quality range from 1 to 5 (worst to best quality). 
The average subjective scores for each image set are reported in Table~\ref{label:user study}.
As summarized in Table~\ref{label:user study}, Zero-DCE achieves the highest average User Study (US) score for a total of 202 testing images from the above-mentioned image sets.
For the MEF, DICM, and VV sets, our results are most favored by the subjects.
%
%
In addition to the US score, we employ a non-reference perceptual index (PI)~\cite{PI,Ma2017,NIQE} to evaluate the perceptual quality. The PI metric is originally used to measure perceptual quality in image super-resolution. It has also been used to assess the performance of other image restoration tasks, such as image dehazing~\cite{Qu2019}. A lower PI value indicates better perceptual quality. The PI values are reported in Table~\ref{label:user study} too.
Similar to the user study, the proposed Zero-DCE is superior to other competing methods in terms of the average PI values.


\vspace{-12pt}
\subsubsection{Quantitative Comparisons}

For full-reference image quality assessment, we employ the Peak Signal-to-Noise Ratio (PSNR,dB), Structural Similarity (SSIM)~\cite{SSIM}, and Mean Absolute Error (MAE) metrics to quantitatively compare the performance of different methods on the Part2 subset~\cite{Cai2018}.  
In Table~\ref{table_1}, the proposed Zero-DCE achieves the best values under all cases, despite that it does not use any paired or unpaired training data.
Zero-DCE is also computationally efficient, benefited from the simple curve mapping form and lightweight network structure.  Table~\ref{table_2} shows the runtime\footnote[4]{Runtime is measured on a PC with an Nvidia GTX 2080Ti GPU and Intel I7 6700 CPU, except for Wang \etal~\cite{Wang2019}, which has to run on GTX 1080Ti GPU.} of different methods averaged on 32 images of size 1200$\times$900$\times$3. For conventional methods, only the codes of CPU version are available.

\begin{table}[t]
	\caption{Quantitative comparisons in terms of full-reference image quality assessment metrics. The best result is in red whereas the second best one is in blue under each case.}
	\vspace{-0.2cm}
	\centering
	\begin{tabular}{c|c|c|c}
		\hline
		\textbf{Method} & \textbf{PSNR$\uparrow$ } & \textbf{SSIM$\uparrow$}  & \textbf{MAE$\downarrow$} \\
		\hline
		SRIE~\cite{Fu2016}           &  14.41      & 0.54 & 127.08\\
		LIME~\cite{Guo2017}          &  16.17      & {\color{blue}0.57} & 108.12\\
		Li \etal~\cite{Li2018}          &  15.19      &   0.54 &114.21         \\
		RetinexNet~\cite{Chen2018}   &  15.99      & 0.53 & 104.81 \\
		Wang \etal~\cite{Wang2019}      &  13.52      &0.49  & 142.01\\
		EnlightenGAN~\cite{Jiang2019}&  {\color{blue}16.21}    & {\color{red}0.59} & {\color{blue}102.78}\\
		Zero-DCE                      &  {\color{red}16.57}     & {\color{red}0.59} & {\color{red}98.78}\\
		\hline
	\end{tabular}
	\vspace{-5pt}
	\label{table_1}
\end{table}

\begin{table}[t]
	\caption{Runtime (RT) comparisons (in second). The best result is in red whereas the second best one is in blue.}
	\vspace{-0.2cm}
	\centering
	\begin{tabular}{c|c|c}
		\hline
		\textbf{Method} & \textbf{RT} & \textbf{Platform} \\
		\hline
		SRIE~\cite{Fu2016}           &  12.1865                            & MATLAB (CPU) \\
		LIME~\cite{Guo2017}          &  0.4914                             & MATLAB (CPU)\\
		Li \etal~\cite{Li2018}       &  90.7859                            & MATLAB (CPU) \\
		RetinexNet~\cite{Chen2018}   &  0.1200                             & TensorFlow (GPU)\\
		Wang \etal~\cite{Wang2019}      &  0.0210                           & TensorFlow (GPU)\\
		EnlightenGAN~\cite{Jiang2019}&  {\color{blue}0.0078}              & PyTorch (GPU)\\
		Zero-DCE                     &  {\color{red}0.0025}                & PyTorch (GPU)\\
		\hline
	\end{tabular}
	\vspace{-8pt}
	\label{table_2}
\end{table}

\vspace{-10pt}
\subsubsection{Face Detection in the Dark}
\begin{figure}[htb]
	\vspace{-0.9cm}
	\centering
	\centerline{\includegraphics[width=0.85\linewidth]{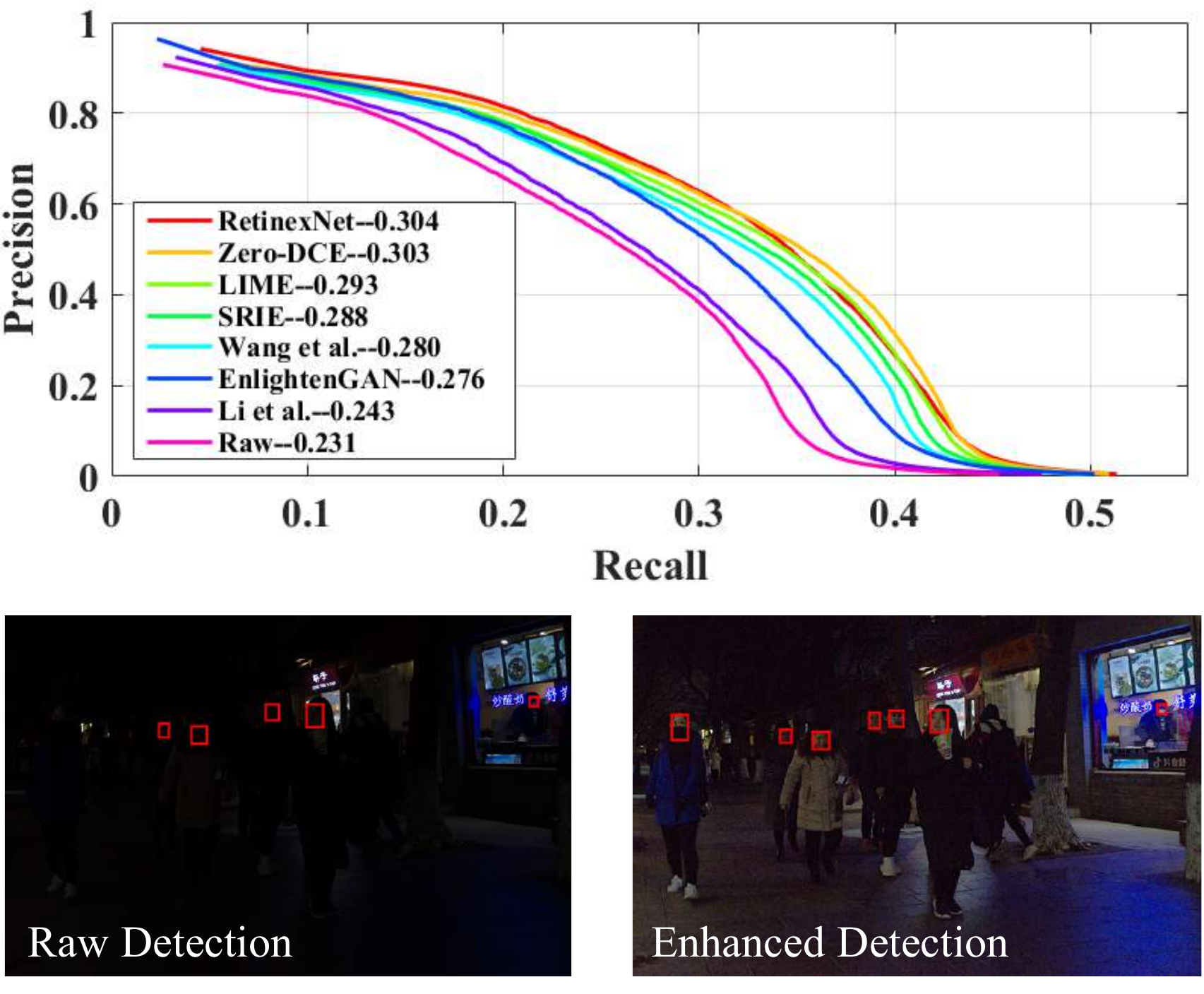}}
		\vspace{-0.3cm}
	\caption{The performance of face detection in the dark. P-R curves, the AP, and two examples of face detection before and after enhanced by our Zero-DCE.}
	\label{fig:PR}
		\vspace{-0.55cm}
\end{figure}

We investigate the performance of low-light image enhancement methods on the face detection task under low-light conditions. Specifically, we use the latest DARK FACE dataset~\cite{2019arXiv190404474Y} that composes of 10,000 images taken in the dark.
Since the bounding boxes of test set are not publicly available, we perform evaluation on the training and validation sets, which consists of 6,000 images.
A state-of-the-art deep face detector, Dual Shot Face Detector (DSFD)~\cite{DSFD}, trained on WIDER FACE dataset~\cite{WIDER}, is used as the baseline model. We feed the results of different low-light image enhancement methods to the DSFD~\cite{DSFD} and depict the precision-recall (P-R) curves in Fig.~\ref{fig:PR}. Besides, we also compare the average precision (AP) by using the evaluation tool\footnote[5]{\url{https://github.com/Ir1d/DARKFACE_eval_tools}} provided in DARK FACE dataset~\cite{2019arXiv190404474Y}.

As shown in Fig.~\ref{fig:PR}, after image enhancement, the precision of DSFD~\cite{DSFD} increases considerably compared to that using raw images without enhancement. Among different methods, RetinexNet~\cite{Chen2018} and Zero-DCE perform the best. Both methods are comparable but Zero-DCE performs better in the high recall area. Observing the examples, our Zero-DCE lightens up the faces in the extremely dark regions and preserves the well-lit regions, thus improves the performance of face detector in the dark.

\vspace{-8pt}
\section{Conclusion}
\vspace{-6pt}
We proposed a deep network for low-light image enhancement. It can be trained end-to-end with zero reference images. This is achieved by formulating the low-light image enhancement task as an image-specific curve estimation problem, and devising a set of differentiable non-reference losses. Experiments demonstrate the superiority of our method against existing light enhancement methods. In future work, we will try to introduce semantic information to solve hard cases and consider the effects of noise.

\vspace{-10pt}
\paragraph{Acknowledgements.}
\footnotesize{This research was supported by NSFC (61771334,61632018,61871342), SenseTime-NTU AI Collaboration Project, Singapore MOE AcRF Tier 1 (2018-T1-002-056), NTU SUG, NTU NAP, Fundamental Research Funds for the Central Universities (2019RC039), China Postdoctoral Science Foundation (2019M660438), Hong Kong RGG (9048123) (CityU 21211518), Hong Kong GRF-RGC General Research Fund (9042322,9042489,9042816).}

{\small
\bibliographystyle{ieee_fullname}
\bibliography{egbib}
}

\end{document}